\documentclass{article}
\usepackage{amssymb}

\usepackage[preprint]{corl_2026} 
\usepackage{microtype}
\usepackage{amsmath, amssymb}
\usepackage{graphicx}
\usepackage{stfloats}
\usepackage{booktabs, multirow}
\usepackage{balance}
\usepackage{float}
\usepackage{wrapfig}
\usepackage{caption}
\usepackage{enumitem}
\newcommand{\bl}[1]{{\flushleft\textbf{#1}}}
\newcommand{\meanstd}[2]{$#1${\scriptsize$\;\pm\,#2$}}
\newcommand{\meanstdtiny}[2]{$#1${\tiny$\;\pm\,#2$}}

\usepackage[table]{xcolor}
\usepackage{nicematrix}
\usepackage[normalem]{ulem}
\usepackage[frozencache,cachedir=minted-main]{minted}

\definecolor{Mahogany}{RGB}{192,64,0}
\definecolor{Bittersweet}{RGB}{254,111,94}
\definecolor{RoyalBlue}{RGB}{65,105,225}
\definecolor{BrickRed}{RGB}{203,65,84}
\definecolor{Goldenrod}{RGB}{218,165,32}
\definecolor{OliveGreen}{RGB}{85,107,47}
\definecolor{Fuchsia}{RGB}{255,0,255}
\definecolor{White}{RGB}{255,255,255}

\definecolor{dusty}{RGB}{228,231,235}   
\definecolor{beige}{RGB}{245,238,230}   
\definecolor{bluegray}{RGB}{235,240,247} 

\definecolor{blue}{HTML}{1F5AA6}
\definecolor{orange}{HTML}{B35A00}
\definecolor{green}{HTML}{1C7A3A}
\definecolor{red}{HTML}{A32323}
\definecolor{purple}{HTML}{6A3D9A}
\definecolor{brown}{HTML}{8C564B}
\definecolor{pink}{HTML}{C04B91}

\newcommand{\blue}[1]{\textcolor{blue}{#1}}
\newcommand{\red}[1]{\textcolor{red}{#1}}
\newcommand{\green}[1]{\textcolor{green}{#1}}
\newcommand{\orange}[1]{\textcolor{orange}{#1}}
\newcommand{\purple}[1]{\textcolor{purple}{#1}}
\newcommand{\pink}[1]{\textcolor{pink}{#1}}
\newcommand{\brown}[1]{\textcolor{brown}{#1}}

\definecolor{AccentBlue}{RGB}{20,90,160}
\newcommand{\mbe}[1]{\textcolor{AccentBlue}{\textbf{#1}}}

\title{RankQ: Offline-to-Online Reinforcement Learning \\ via Self-Supervised Action Ranking}

\author{
    Andrew Choi and Wei Xu$^{\dagger}$ \\
    Horizon Robotics \ $^\dagger$Corresponding author \\
    \texttt{\footnotesize https://horizonrobotics.github.io/gail/rankq}
}

\begin{document}
\maketitle


\vspace{-1.00cm}
\begin{abstract}
Offline-to-online reinforcement learning (RL) improves sample efficiency by leveraging pre-collected datasets prior to online interaction.
A key challenge, however, is learning an accurate critic in large state--action spaces with limited dataset coverage.
To mitigate harmful updates from value overestimation, prior methods impose pessimism by down-weighting out-of-distribution (OOD) actions relative to dataset actions.
While effective, this essentially acts as a behavior cloning anchor and can hinder downstream online policy improvement when dataset actions are suboptimal.
We propose RankQ, an offline-to-online Q-learning objective that augments temporal-difference learning with a self-supervised multi-term ranking loss to enforce structured action ordering.
By learning relative action preferences rather than uniformly penalizing unseen actions, RankQ shapes the Q-function such that action gradients are directed toward higher-quality behaviors.
Across sparse-reward D4RL benchmarks, RankQ achieves competitive overall performance against seven baselines.
In vision-based robot learning, RankQ enables effective offline-to-online fine-tuning of a pretrained vision-language-action (VLA) model in a low-data regime, achieving an average simulation success rate 38.2 percentage points higher than the next best method.
In a high-data setting, RankQ improves simulation performance by 13.7 percentage points over the next best method and demonstrates sim-to-real transfer, increasing real-world cube stacking success from 43.1\% to 88.9\% relative to the VLA’s initial performance.
Video results are available at our \href{https://horizonrobotics.github.io/gail/rankq}{project page}.

\end{abstract}

\vspace{-0.25cm}

\section{Introduction}
Reinforcement learning (RL) has achieved strong results across many domains, but its reliance on extensive online interaction remains a key limitation.
In real-world robotics, where data collection is expensive or potentially unsafe, this challenge is further exacerbated.
Offline RL~\citep{levine2020offlinereinforcementlearningtutorial} addresses this issue by learning entirely from static, pre-collected datasets, avoiding additional environment interaction during training.
Building upon this paradigm, offline-to-online RL combines offline pretraining with online fine-tuning to improve sample efficiency while still enabling policy improvement beyond the offline dataset.
However, inaccurate value estimates during offline pretraining can lead to harmful policy updates during subsequent online fine-tuning, particularly for OOD actions.
A common solution is to enforce pessimism over unseen actions~\citep{nakamoto2023calql, kumar2020conservativeqlearningofflinereinforcement}, effectively biasing the policy toward dataset actions.
While this stabilizes training, it can be overly restrictive, hindering downstream online improvement when dataset actions are suboptimal.
This raises a key question: how should value functions be shaped beyond the dataset to enable effective online policy improvement?

We propose RankQ, an offline-to-online Q-learning objective that augments the temporal-difference (TD) loss with a self-supervised ranking objective.
Rather than uniformly penalizing unseen actions, RankQ models the relative quality of actions and enforces consistent ordering between high- and low-quality behaviors during both offline and online learning.
This directly shapes the Q-value landscape such that gradients with respect to actions, $\partial Q / \partial a$, are directed toward regions containing higher-quality actions.
As a result, RankQ enables more effective policy improvement throughout offline pretraining and online fine-tuning.
Overall, our main contributions are as follows:
\begin{enumerate}[leftmargin=12pt, topsep=0pt, itemsep=0pt]
\item \textbf{Strong performance on offline-to-online RL benchmarks.}
We evaluate RankQ against seven baselines on four $\texttt{antmaze}$ and three $\texttt{adroit}$ sparse-reward environments from the D4RL benchmark~\citep{fu2021d4rldatasetsdeepdatadriven}, where it \mbe{achieves competitive overall performance}.
\item \textbf{Effective VLA offline-to-online RL fine-tuning in the low-data regime.}
Unlike models trained from scratch, pretrained vision-language-action (VLA) models can achieve strong initial performance on in-distribution tasks, enabling RL fine-tuning with only sparse rewards~\citep{li2026simplevlarl}.
We fine-tune VLAs in simulation under a challenging pseudo real-data regime, consisting of offline pretraining on self-collected rollouts followed by limited online interaction.
In this setting, \mbe{RankQ is the only method that consistently and substantially improves over the baseline, achieving an average success rate 38.2 percentage points higher than the next best method across three tasks}.
In this data regime, we also provide an in-depth ablation studying the key components of RankQ.
\item \textbf{Effective VLA offline-to-online RL fine-tuning in the high-data regime and sim-to-real validation.}
In a high-data setting with large-scale simulation and domain randomization, \mbe{RankQ improves success rate by 13.7 percentage points and reduces task completion time by 20.5\% over the next best method}.
We further demonstrate sim-to-real transfer across 288 real-world trials, where a RankQ-trained policy \mbe{improves cube stacking success from 43.1\% to 88.9\%}.
\item \textbf{Extensive analysis of RankQ.}
We study the effects of RankQ on the Q-value landscape through an extensive set of analyses.
We provide additional ablations, Q-landscape statistics, and critic accuracy comparisons against baseline methods in Secs.~\ref{sec:rankq_ablation}, \ref{sec:q-landscape-analysis}, and~\ref{sec:critic-accuracies}, respectively.
\end{enumerate}

\section{Related Work}

Offline RL learns policies purely from static datasets without additional environment interaction, while offline-to-online RL initializes policies from offline datasets and subsequently improves them through online interaction~\citep{levine2020offlinereinforcementlearningtutorial, lee2021offlinetoonlinereinforcementlearningbalanced}.
A major challenge in both settings is preventing harmful policy updates caused by inaccurate value estimates, particularly for OOD states and actions.
Such issues become especially severe under low-coverage or imbalanced datasets~\citep{hong2023beyond}.

A large class of offline RL methods mitigates extrapolation error by constraining policies toward the support of the offline dataset.
Methods such as BCQ~\citep{fujimoto2019offpolicydeepreinforcementlearning}, BEAR~\citep{kumar2019stabilizingoffpolicyqlearningbootstrapping}, and BRAC~\citep{wu2019behaviorregularizedofflinereinforcement} explicitly constrain policy updates toward dataset actions.
Subsequent methods instead combine RL objectives with behavior cloning regularization.
AWAC~\citep{nair2021awacacceleratingonlinereinforcement}, TD3+BC~\citep{beeson2022improvingtd3bcrelaxedpolicy}, and ReBRAC~\citep{tarasov2023revisitingminimalistapproachoffline} stabilize learning through various forms of behavior regularization, while IQL~\citep{kostrikov2022offline} avoids explicit optimization over OOD actions using expectile regression and advantage-weighted updates~\citep{peng2019advantageweightedregressionsimplescalable}.
While originally proposed for offline RL, many of these methods are also commonly used to initialize policies for subsequent online fine-tuning.

Offline-to-online RL methods additionally study how offline and online data should be combined during fine-tuning.
Hybrid RL~\citep{song2023hybrid} uses an offline training stage and afterwards combines offline and online replay buffers to improve sample efficiency, while RLPD~\citep{ball2023efficientonlinereinforcementlearning} shows that simply incorporating offline data into standard off-policy RL can greatly accelerate online learning.
WSRL~\citep{zhou2025wsrl} instead separates the two phases, using offline RL to initialize learning before discarding the offline data and warm-starting standard online RL with newly collected experience.
Other approaches instead modify the training architecture or data pipeline.
Ensemble-based methods~\citep{zhao2024qensemble, lee2021offlinetoonlinereinforcementlearningbalanced, an2021uncertaintybasedofflinereinforcementlearning} improve value estimation through critic ensembles, while Policy Expansion (PEX)~\citep{zhang2023policy} freezes the offline policy and trains a second online policy during fine-tuning to maintain the skills obtained by the offline policy.
Sequence modeling approaches such as Online Decision Transformer (ODT)~\citep{zheng2022odt} extend Decision Transformer~\citep{chen2021decisiontransformerreinforcementlearning} to the offline-to-online RL setting, while more recent approaches such as CFDG~\citep{huang2025cfg} use diffusion models to generate synthetic offline and online experiences.

A prominent line of work in both offline RL and offline-to-online RL instead enforces pessimism toward unseen actions.
Conservative Q-Learning (CQL)~\citep{kumar2020conservativeqlearningofflinereinforcement} penalizes OOD actions by lowering their Q-values with respect to dataset actions, while Calibrated Q-Learning (Cal-QL)~\citep{nakamoto2023calql} improves upon CQL's potential over-pessimism by calibrating value estimates with respect to a reference policy to stabilize online fine-tuning.
Adaptive Policy Learning (APL)~\citep{zheng2023apl} further applies conservative updates only to offline transitions while performing optimistic updates on online data.
Such methods achieve strong performance and stable fine-tuning, and have been used for chunked RL~\citep{huang2025corftefficientfinetuningvisionlanguageaction} and even enabled real-world robotic RL fine-tuning within as little as one hour~\citep{chen2025conrft}.
However, these approaches fundamentally assume that all unseen actions are equally less preferable than dataset actions, which can hinder policy improvement when the offline dataset contains suboptimal behavior~\citep{hong2023beyond}.
In contrast, RankQ explicitly models relative action quality through structured ranking objectives.
Rather than uniformly penalizing unseen actions, RankQ shapes the value landscape using self-supervised action comparisons, encouraging action gradients to consistently point toward higher-quality regions of the action space.

\begin{figure}[t]
\includegraphics[width=\columnwidth]{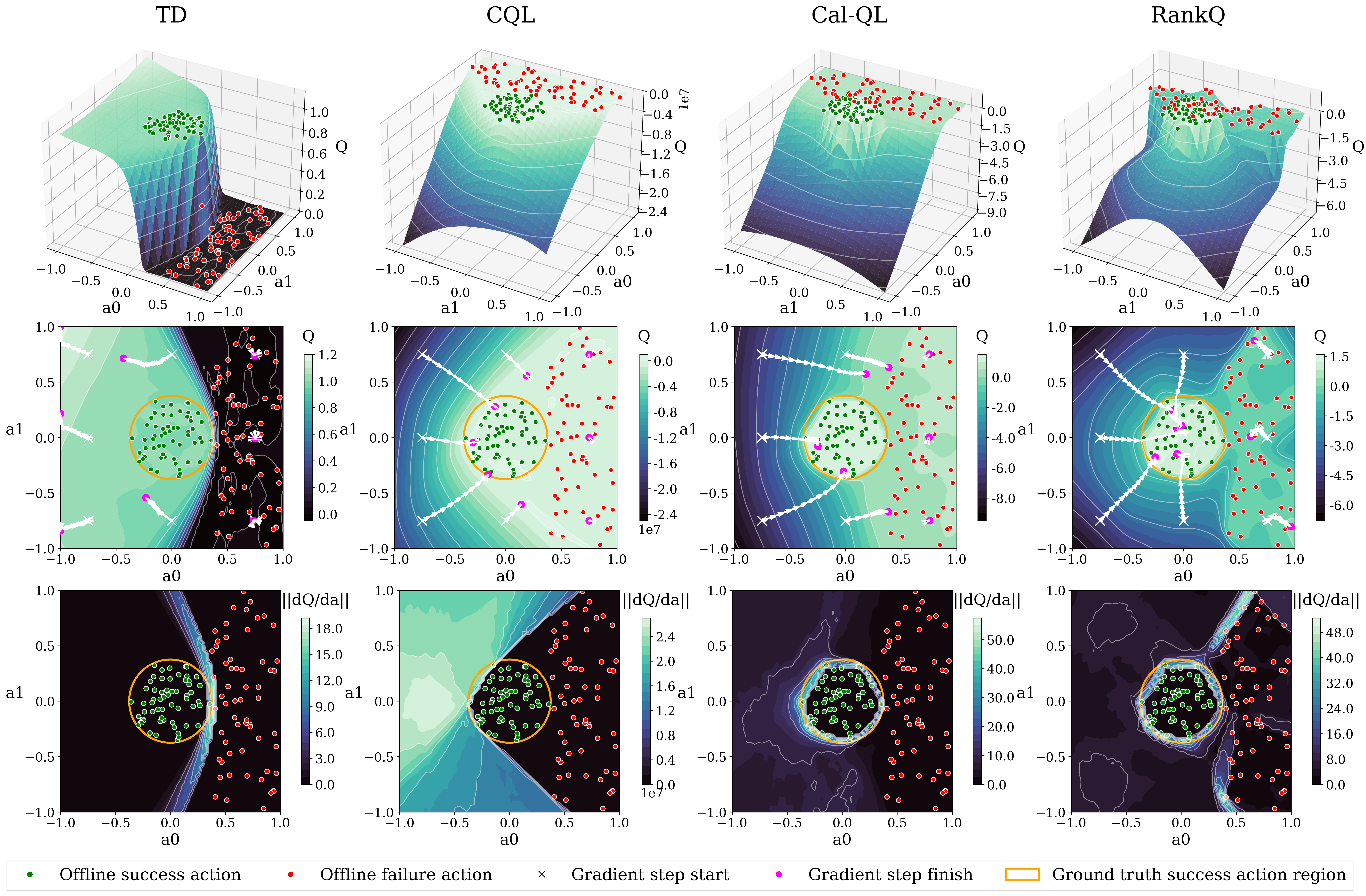}
\caption{
Toy example visualizing the Q-landscape for a fixed state $s$ and 2D action space $(a_0, a_1) \in \mathcal A$ under different critic objectives after offline training.
The first row shows the 3D Q-manifold; the second row shows a 2D projection with eight sampled actions and their gradient ascent trajectories; and the third row shows $\partial Q/\partial a$ magnitudes.
By explicitly shaping the Q-landscape, RankQ enables the largest number of trajectories to converge toward the optimal action region.
}
\label{fig:toy_example}
\end{figure}


\section{Methodology}
For a Markov decision process with state space $\mathcal{S}$, action space $\mathcal{A}$, reward function $r$, and discount factor $\gamma \in [0, 1]$, the goal of RL is to learn a policy $\pi(a|s)$ that maximizes the expected discounted return $J(\pi) = \mathbb{E}_\pi[ \sum_{t=0}^{T-1} \gamma^t r_t ]$, where $T$ is the episode length.
In actor-critic methods, a critic network is trained to estimate the action-value function $Q^\pi(s,a) = \mathbb{E}_{s'} \left[ r(s,a) + \gamma V^\pi(s') \right]$, which provides gradients to update the actor toward maximizing the discounted return.
To train the critic, temporal-difference (TD) learning~\citep{sutton2018reinforcement} is used, which regresses current value estimates toward bootstrapped targets. This is expressed as
\begin{equation}
\mathcal{L}_{\textrm{TD}}(\theta) = \frac{1}{2} \mathbb{E}_{s,a,s' \sim \mathcal{D}} \left[
\left(
Q_\theta(s,a) - \left(r(s,a) + \gamma \mathbb{E}_{a' \sim \pi(\cdot \mid s')} \left[ Q_{\bar{\theta}}(s', a') \right]\right)
\right)^2
\right],
\end{equation}
where $\mathcal{D}$ is the replay buffer (or offline dataset in the offline RL setting) and $\bar{\theta}$ denotes the parameters of a lagged target critic used for stability~\citep{mnih2015dqn}.
For the remainder of the paper, we assume $r(s,a)$ to be a sparse reward administered upon rule-based success detection.
In the context of offline RL, naively using TD learning on a static dataset without sufficient coverage can lead to overestimated value estimates for OOD actions (first column of Fig.~\ref{fig:toy_example}).
This value overestimation can then lead to poorly informed updates to the actor model, which can hinder or crash training entirely.

\bl{Conservative Q-Learning.}
To address value overestimation, Conservative Q-Learning (CQL)~\citep{kumar2020conservativeqlearningofflinereinforcement} explicitly penalizes Q-values of actions sampled from the policy, encouraging them to be lower than those observed in the dataset, by augmenting the TD loss with a conservative regularizer:
\begin{equation}
    \mathcal L^{\textrm{CQL}}_Q(\theta) =  \blue{\alpha \left(\mathbb E_{s\sim\mathcal D, a \sim \pi(\cdot | s)} [ Q_\theta(s, a)] - \mathbb E_{s,a\sim \mathcal D} [ Q_\theta(s,a)] \right)} + \mathcal L_\textrm{TD}(\theta),
\label{eq:cql}
\end{equation}
where $\alpha$ is a hyperparameter controlling the strength of the regularizer. 
This effectively imposes pessimism over OOD actions, biasing the learned policy toward the data distribution. 
While effective at preventing overestimation, this objective has two key issues.
First, the pessimism on OOD actions is unbounded, which with enough offline training, can lead to large Q-value drops resulting in excessively large $\partial Q/\partial a$ gradients.
Observe how in the second column of Fig.~\ref{fig:toy_example}, for the same iterations of training, CQL possesses values and gradients several orders of magnitude higher than other methods.
Actor updates with such gradients can cause policy instability and once proper coverage is achieved for those OOD actions via online exploration, a significant number of training updates is usually necessary to essentially unlearn this pessimism~\citep{nakamoto2023calql}.
The second issue of CQL lies in uniform preference of dataset actions to policy ones.
Such treatment does not take into consideration the chance that policy actions may actually be optimal (should have higher $Q(s,a)$) while dataset actions are suboptimal (should have lower $Q(s,a)$).
This can be especially problematic if the offline dataset contains a large portion of failure rollouts.

\bl{Calibrated Q-Learning.}
Building directly off CQL, Calibrated Q-Learning (Cal-QL)~\citep{nakamoto2023calql} solves CQL's issue of unbounded pessimism through ``calibration", i.e., ensuring that policy action Q-values are not pushed down past a reference policy's value $V^\mu(s)$. 
With the following modification,
\begin{equation}
    \mathcal L^{\textrm{Cal-QL}}_Q(\theta) =  \blue{\alpha \left(\mathbb E_{s\sim\mathcal D, a \sim \pi(\cdot | s)} [ \red{ \max (Q_\theta(s, a), V^\mu(s)) }] - \mathbb E_{s,a\sim \mathcal D} [ Q_\theta(s,a)] \right)} + \mathcal L_\textrm{TD}(\theta),
\label{eq:calql}
\end{equation}
Cal-QL prevents excessive suppression of potentially good actions while still mitigating overestimation. 
Indeed, upon inspecting the third column of Fig.~\ref{fig:toy_example}, we can observe that the scale of OOD Q-values is much more reasonable compared to CQL.
Regardless of the calibrated floor of Q-values, Cal-QL and CQL still suffer from their uniform preference of dataset actions over others.
This reflects directly in the $\partial Q / \partial a$ gradient flow field where many sampled actions have a flow that leads them to dataset failure actions rather than dataset success actions.

\bl{RankQ.}
Rather than treating all OOD actions as equally inferior, RankQ enforces structured action ordering in a way that shapes the Q-landscape to point toward optimal actions.  
We partition the dataset $\mathcal D$ into two unique subsets containing exclusively success and failure trajectories, denoted by $\mathcal D_{\text{success}}$ and $\mathcal D_{\text{failure}}$.
For transitions $(s,a) \sim \mathcal D_{\text{success}}$, we construct several classes of self-supervised suboptimal actions that capture different types of deviations from successful behavior:
\begin{itemize}[leftmargin=12pt, topsep=0pt, itemsep=0pt]
    \item \textbf{Noisy actions:} $a_\epsilon = a + \epsilon$, where $\epsilon \sim \mathcal N(0, \sigma^2)$, representing actions with small perturbations.
    \item \textbf{Very noisy actions:} $a_{2\epsilon} = a + 2\epsilon$, representing actions with larger perturbations.
    \item \textbf{Random actions:} $a_r \sim \mathcal U(-1,1)^{|a|}$, representing arbitrary actions.
    \item \textbf{Permuted actions:} $a_p \sim \mathcal D$, representing actions sampled from unrelated states.
\end{itemize}

These constructions allow us to impose structured ordering constraints. First, we enforce that successful actions are preferred over all suboptimal variants:
\begin{equation}
Q(s,a) > Q(s,a'), \quad
a' \in \{a_{\epsilon}, a_{2\epsilon}, a_r, a_p\}, \;
(s,a) \sim \mathcal{D}_{\mathrm{success}}.
\label{eq:success_ranks}
\end{equation}
Enforcing only Eq.~\ref{eq:success_ranks} would essentially produce a Q-landscape with a gradient field similar to CQL and Cal-QL.
Rather than stopping here, we also enforce ordering among suboptimal actions by using action-space proximity to successful actions as a heuristic for relative quality:
\begin{equation}
Q(s,a_{\epsilon}) > Q(s,a_{2\epsilon}) > Q(s,a_r).
\label{eq:success_neg_ranks}
\end{equation}
This proximity-based ordering is a heuristic and may not hold for every individual action. 
However, the ranking constraints are imposed in expectation, and the softplus loss becomes small once an ordering is satisfied, limiting continued pressure from potentially incorrect rankings. 
Moreover, TD supervision remains active throughout training and can correct inconsistent value estimates.
Finally, we incorporate failure trajectories to provide additional supervision in low-quality regions of the state space. We enforce that a failure dataset action is still preferable to random actions:
\begin{equation}
Q(s,a) > Q(s,a_r), \; (s,a)\sim \mathcal D_\textrm{failure},
\label{eq:failure_ranks}
\end{equation}
but refrain from imposing stronger ordering constraints due to the absence of a meaningful notion of relative quality.
The full multi-term pairwise ranking loss can be seen formulated in Sec.~\ref{sec:rankq_loss}.

\section{Experiments}

\bl{Environments.}
We evaluate RankQ across the following environments and data regimes:
\begin{enumerate}[leftmargin=12pt, topsep=0pt, itemsep=0pt]
    \item \texttt{antmaze-medium}: An 8-DOF ant must navigate to a goal location within a maze.
    \item \texttt{antmaze-large}: A more difficult version of $\texttt{antmaze-medium}$ with a larger maze.
    \item \texttt{adroit-pen}: A dexterous 24-DOF hand must manipulate a pen to reach a desired orientation.
    \item \texttt{adroit-door}: Same as above, but policy must open a door.
    \item \texttt{adroit-relocate}: Same as above, but policy must grab a ball and move it to a specified position.
    \item \texttt{vla-low-data-carrot-onto-plate}: A VLA is fine-tuned in a low-data regime to control a 7-DOF manipulator using action chunks of size 4 (total 28 DOFs) to place a carrot onto a plate.
    \item \texttt{vla-low-data-spoon-into-bowl}: Same as above, but policy must place a spoon inside a bowl.
    \item \texttt{vla-low-data-cube-stacking}: Same as above, but policy must stack cubes.
    \item \texttt{vla-sim2real-cube-stacking}: Same as above, but the VLA is trained in a high-data regime along with extensive domain randomization. 
\end{enumerate}
For each $\texttt{antmaze}$ environment, we also evaluate using two types of offline datasets: the $\texttt{play}$ set (goal-oriented demonstrations) and the more difficult $\texttt{diverse}$ set (a combination of random, scripted, and goal-oriented demonstrations).
To keep comparisons as fair as possible, we directly reuse the Cal-QL codebase of~\citet{nakamoto2023calql} with their tuned parameters for the D4RL evaluation.
For all $\texttt{vla-low-data}$ environments, we collect 200 VLA self-rollouts to use as the offline dataset while $\texttt{vla-sim2real}$ uses 800 self-rollouts.

\bl{Baselines.}
For the D4RL benchmarks, we use Soft Actor-Critic (SAC)~\citep{haarnoja2017soft} as the base RL algorithm with simple MLPs as the critic and actor architectures.
For the VLA benchmarks, we use an off-policy formulation of the PPOFlow~\citep{zhang2025reinflow} algorithm implementation from~\citep{choi2026scalingsimtorealreinforcementlearning} that we denote as SACFlow (full details in Sec.~\ref{sec:sac_flow}).
We use an imitation flow-matching model $\pi_0$~\citep{black2024pi0visionlanguageactionflowmodel} pre-trained on the BridgeV2 dataset~\citep{walke2023bridgedata} as our VLA model.
For baselines, we compare \green{RankQ} against the following offline-to-online training strategies (for VLA-related environments, replace SAC with SACFlow):
\begin{enumerate}[leftmargin=12pt, topsep=0pt, itemsep=0pt]
    \item \blue{CQL}~\citep{kumar2020conservativeqlearningofflinereinforcement}: Uses conservative Q-learning all throughout offline-to-online training.
    \item \blue{CQL+SAC}: Uses CQL during offline training and then switches to regular SAC for online.
    \item \red{Cal-QL}~\citep{nakamoto2023calql}: Uses calibrated Q-learning all throughout offline-to-online training.
    \item \red{Cal-QL+SAC}: Uses Cal-QL during offline training and then switches to regular SAC for online.
    \item \orange{SAC+OFF}: Runs regular SAC on the offline dataset before continuing to online training. The offline dataset is used to soft-start the replay buffer.
    \item \orange{SAC}~\citep{haarnoja2017soft}: No offline pretraining; pure online SAC only.
    \item \purple{Hybrid RL}~\citep{song2023hybrid}: Similar to \orange{SAC+OFF} except the offline dataset is kept separate from the online replay buffer with a 50/50 mix being sampled throughout.
    \item \pink{RLPD}~\citep{ball2023efficientonlinereinforcementlearning}: No offline pretraining; online SAC with a fixed offline buffer and 50/50 sampling mix, critic layer norm, and a 4-critic ensemble.
    \item \brown{WSRL}~\citep{zhou2025wsrl}: Uses Cal-QL during offline training, discards the offline data and warm starts SAC with fresh data. 
    \item \green{RankQ+SAC}: Uses RankQ during offline training and then switches to regular SAC for online.
\end{enumerate}

\bl{D4RL evaluation.}
\begin{figure}[t]
\includegraphics[width=\columnwidth]{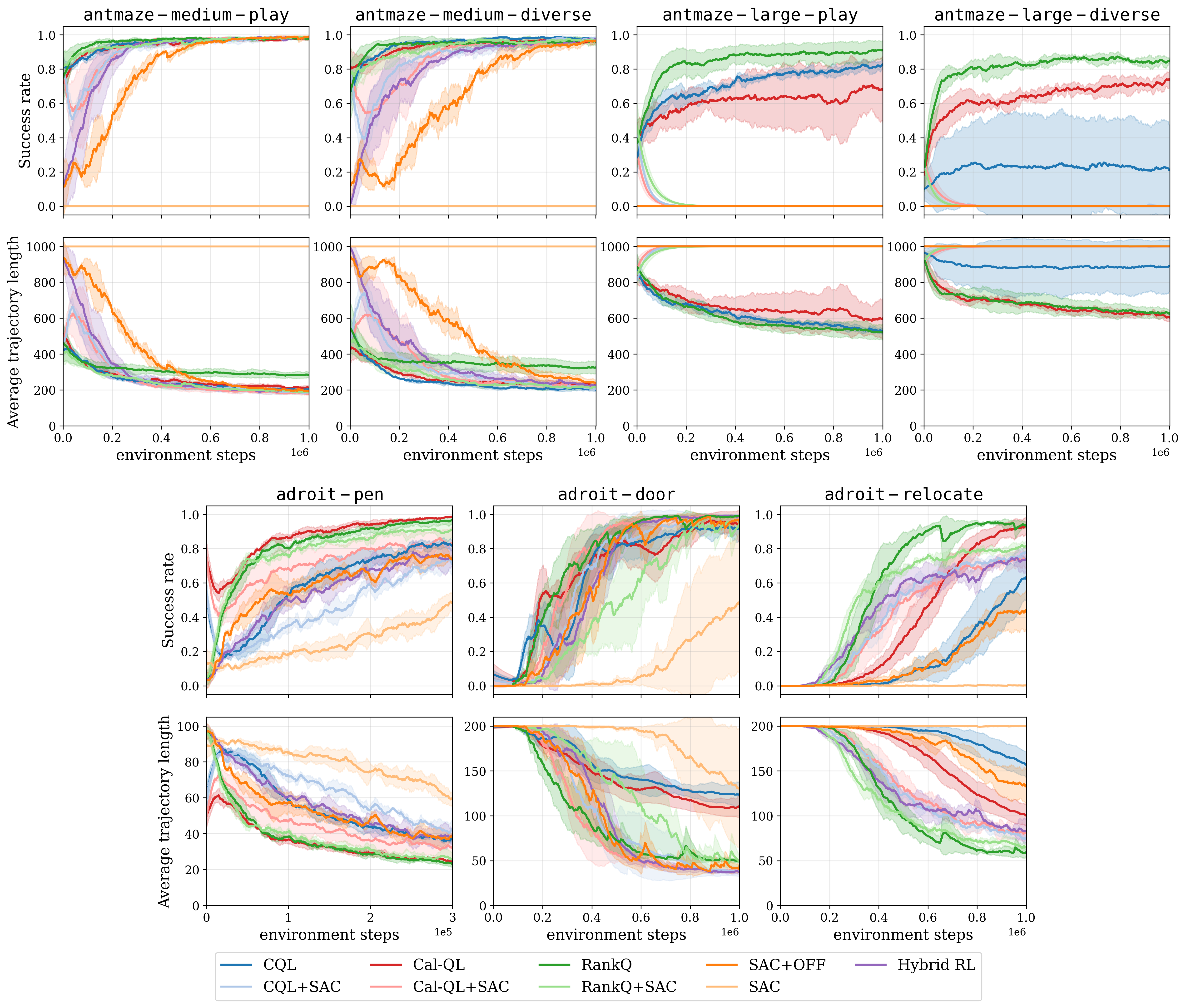}
\caption{Success rate and average trajectory length results for the D4RL \texttt{antmaze} and \texttt{adroit} environments\protect\footnotemark. Curves start after offline RL training has concluded. Each algorithm is reported across 3 random seeds. Note that the average trajectory length metric is failure-inclusive.}
\label{fig:d4rl_results}
\end{figure}
Results on the D4RL benchmarks (Fig.~\ref{fig:d4rl_results}) highlight clear differences in how effectively each method leverages offline data for downstream online improvement.
First, pure online RL (SAC) either learns inefficiently or, in the case of \texttt{antmaze}, fails entirely due to the long-horizon nature of the task.
Naive offline-to-online baselines such as SAC+OFF and Hybrid RL provide substantial improvements over pure online RL, demonstrating the benefit of incorporating offline data.
However, these approaches remain insufficient for more challenging long-horizon tasks and fail to learn meaningful policies on \texttt{antmaze-large}.
The SAC fine-tuning variants of offline-pretrained methods (CQL+SAC, Cal-QL+SAC, and RankQ+SAC) show mixed performance across environments.
While these approaches improve over their full offline-to-online counterparts in some cases, all still fail on \texttt{antmaze-large}, suggesting that switching to standard online RL may fail to preserve benefits acquired during offline pretraining.

Among the full offline-to-online methods, CQL, Cal-QL, and RankQ are the only approaches that achieve non-zero success across all evaluated environments.
RankQ achieves strong overall performance, remaining competitive with the strongest baselines on \texttt{antmaze-medium} and the \texttt{adroit} tasks while providing its clearest gains on the more challenging \texttt{antmaze-large} environments.
In addition to strong success rates, RankQ also produces competitive policies in terms of trajectory efficiency, although this varies by environment.
For example, it exhibits slightly slower trajectories on \texttt{antmaze-medium}, while achieving substantially faster execution on \texttt{adroit-relocate}.
\footnotetext{Some results do not match those reported in~\citep{nakamoto2023calql}. Refer to Sec.~\ref{sec:training_setup} for details.}

\bl{VLA low-data evaluation.}
\begin{figure}[t]
\centering
\includegraphics[width=\columnwidth]{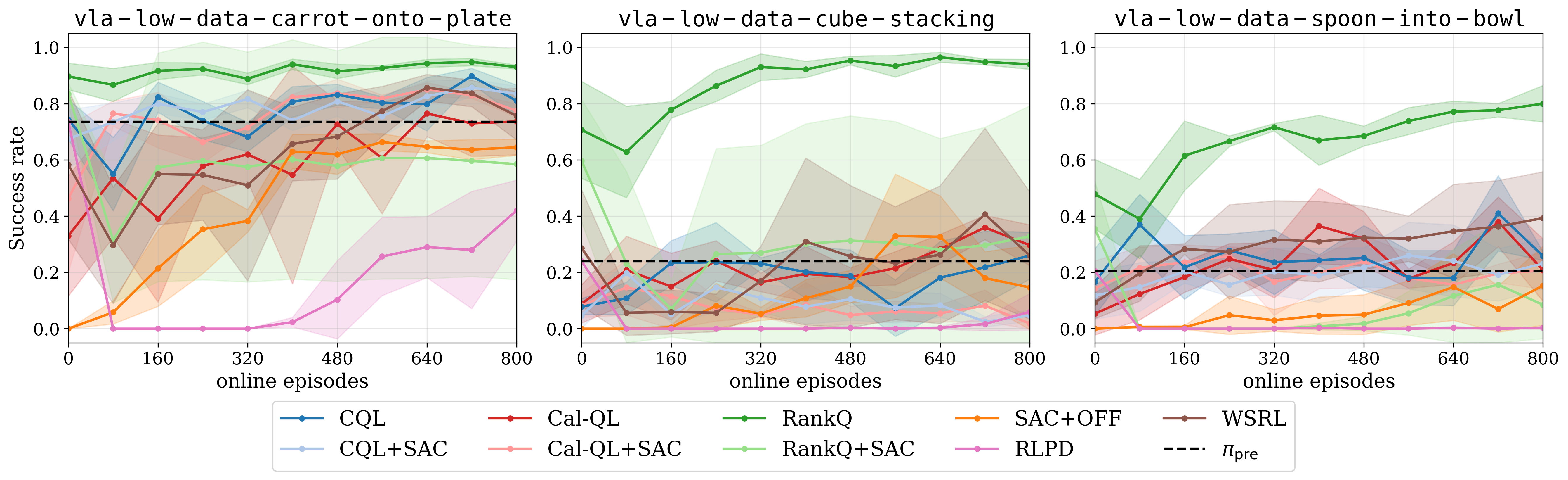}
\caption{Success rate results for \texttt{vla-low-data} environments. 
Curves start after offline RL training has concluded. 
Each algorithm is reported across 3 random seeds with each random seed having its own unique set of 200 self-rollouts.
With only 8 online rollouts per update, RankQ achieves the strongest and most consistent improvement beyond the initial VLA performance across all three tasks.
Though success rate increases, the average time-to-finish was more-or-less flat given the low data budget and is therefore not included.
}
\label{fig:vla-low-data_results}
\end{figure}
We consider a low-data offline-to-online fine-tuning setting for VLA policies, where both offline coverage and online interaction are severely limited.
The offline dataset consists of only 200 self-collected rollouts, and during online training we perform just 8 new rollouts per update.
We model the problem as an infinite-horizon RL task (i.e., episodes do not terminate upon success) so that post-success behavior remains well-defined.
For training efficiency, episodes are truncated after 25 steps via timeout.
We evaluate three manipulation tasks of increasing difficulty: placing a carrot onto a plate, stacking cubes, and placing a spoon into a bowl without tipping it over.
The latter two tasks are particularly challenging, with initial success rates of roughly 20\%, implying that 80\% of the offline dataset consists of failure transitions.
This creates a difficult learning regime in which useful supervision is sparse and naive reliance on dataset actions can be misleading.
Results are shown in Fig.~\ref{fig:vla-low-data_results}, while visualizations of each task are provided in Fig.~\ref{fig:low_data_scenes}.

Under these constraints, RankQ is the only method that consistently improves the policy substantially beyond its imitation baseline.
RLPD performs poorly across all three tasks, suggesting that it is particularly susceptible to low-quality offline data.
WSRL also underperforms despite receiving an additional 200-episode warm-start budget following offline pretraining.
While the remaining approaches show limited or inconsistent improvement, RankQ continues to make steady progress despite the limited budget of only 800 online rollouts.
Interestingly, RankQ+SAC also degrades substantially during online fine-tuning, further indicating that continued ranking-based shaping is important rather than serving only as an offline initialization.
This yields performance improvements over the next best non-RankQ baseline of 9.5, 64.3, and 40.7 percentage points for \texttt{carrot-on-plate}, \texttt{cube-stacking}, and \texttt{spoon-into-bowl}, respectively.
These results demonstrate that leveraging structured action comparisons provides a strong advantage for extracting learning signal from both 
sparse successes and abundant failures, enabling effective policy improvement even in extremely data-constrained settings.
\begin{wrapfigure}{r}{0.45\columnwidth}
    \centering
    \includegraphics[width=\linewidth]{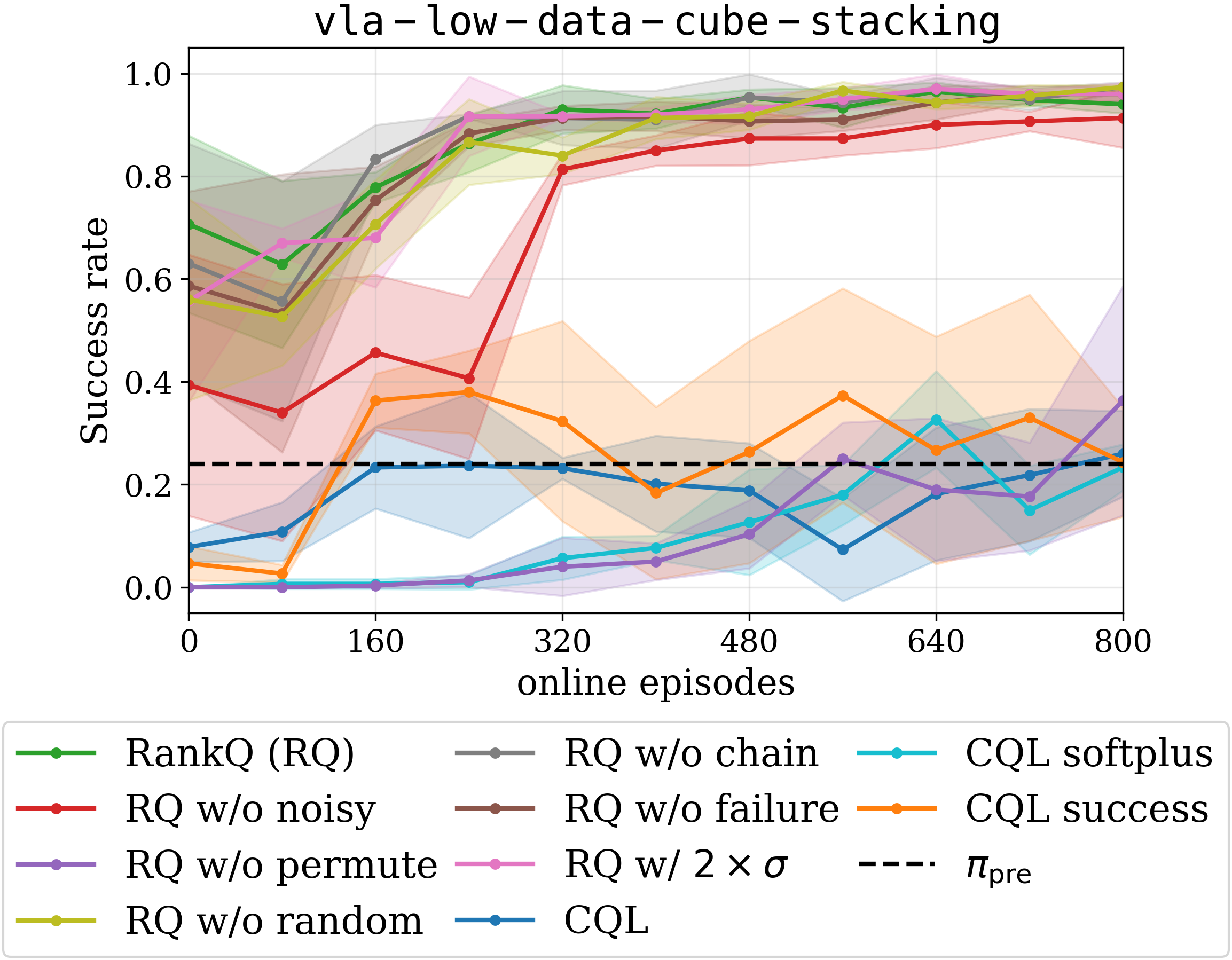}
    \caption{RankQ ablation results.}
    \label{fig:vla_rankq_ablation}
    \vspace{-60pt}
\end{wrapfigure}
\bl{RankQ ablation.}
To better understand which components of RankQ drive its performance gains, we conduct an ablation study on the challenging \texttt{vla-low-data-cube-stacking} environment (Fig.~\ref{fig:vla_rankq_ablation}).
In addition to removing individual ranking terms, we evaluate softplus CQL and success-conditioned CQL as controls to test whether RankQ's gains can be explained simply by smoother conservative regularization or conditioning on successful trajectories.
Both controls remain near the initial VLA performance, suggesting that RankQ's improvement arises from its structured synthetic-action rankings rather than these simpler alternatives.
Ablating individual ranking terms further highlights the importance of this structure.
Removing noisy-action rankings degrades performance, while removing permuted-action rankings causes a substantial collapse, indicating that both local action perturbations and comparisons against actions from unrelated states provide important supervision.
In contrast, removing random-action rankings has little effect on performance.
We hypothesize that this is because the pretrained VLA already strongly biases actions away from random behaviors.
Still, their inclusion does not degrade performance and provides additional shaping in poorly covered regions.

\bl{VLA high-data evaluation.}
\begin{figure}[t]
\centering
\includegraphics[width=0.8\columnwidth]{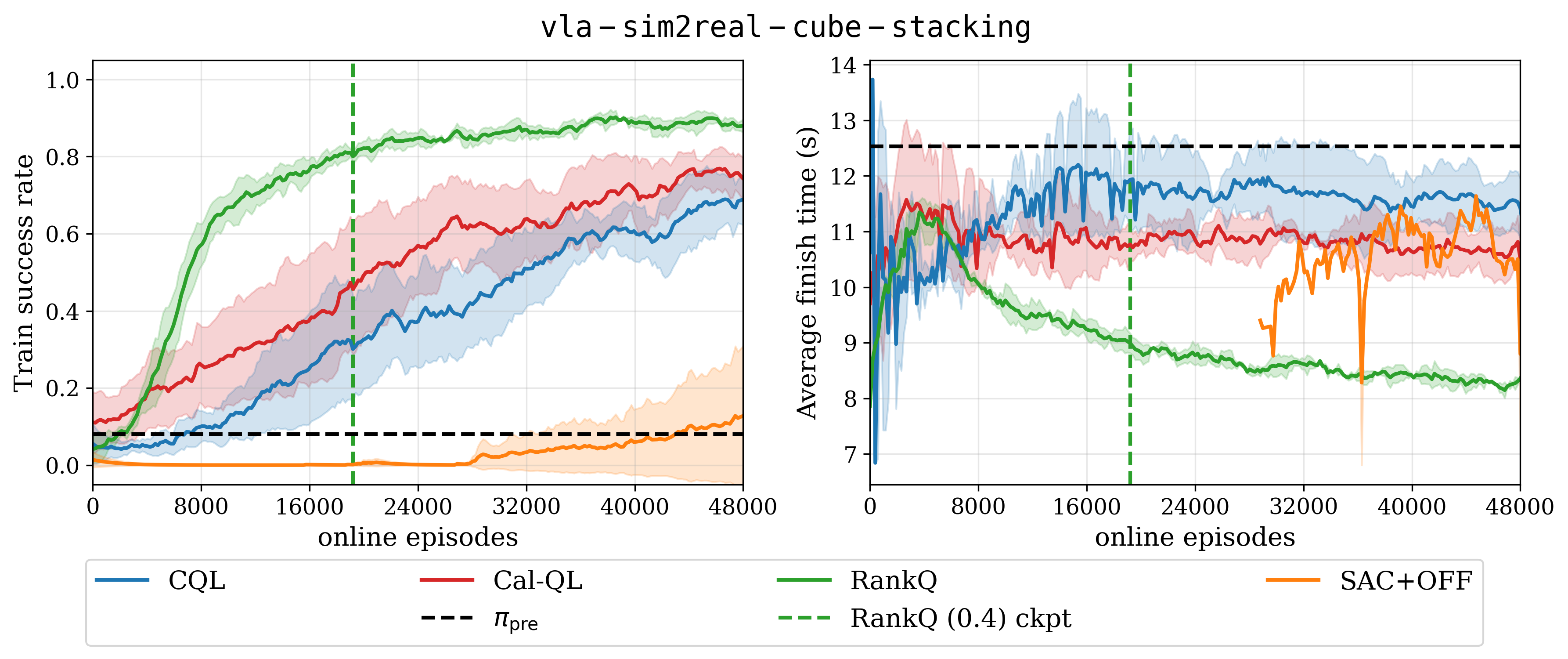}
\caption{
Training rollout success rate and average time-to-finish results for the~\texttt{vla-sim2real-} \texttt{cube-stacking} environment.
Note that training rollout metrics are slightly lower than evaluation metrics due to stochastic action sampling rather than greedy mode selection.
Curves begin after offline RL training has concluded.
Each algorithm is reported across 3 random seeds, with each seed using a unique set of 800 self-rollouts.
The average finish time is failure-exclusive.
}
\label{fig:vla-sim2real_results}
\end{figure}
We evaluate a higher-data regime on the cube stacking task (Fig.~\ref{fig:vla-sim2real_results}), with the objective of learning a policy that zero-shot transfers from simulation to the real world.
The training setup follows the same formulation as the low-data setting, but is scaled to improve coverage and robustness.
Different from the low-data evaluation, we apply extensive domain randomization (Table~\ref{tab:domain_randomization}) in the form of scene, robot pose, camera pose, and lighting variations.
Visualizations of the randomized scenes are shown in Fig.~\ref{fig:high_data_scenes}.
The offline dataset is increased to 800 self-rollouts. However, with an initial success rate of approximately 8\%, this still corresponds to a highly imbalanced dataset containing only about 64 successful trajectories.
During online fine-tuning, we additionally increase the number of rollouts per update to 192, providing substantially more interaction compared to the low-data regime.

With this increased data budget, pessimism-based methods such as CQL and Cal-QL are able to improve beyond the imitation baseline (Fig.~\ref{fig:vla-sim2real_results}), unlike in the low-data setting where they largely fail to make progress.
However, their sample efficiency remains significantly lower than RankQ, which improves more rapidly and achieves a final success rate 13.7 percentage points higher than Cal-QL.
In contrast, most random seeds of naive offline-to-online approaches such as SAC+OFF still fail to learn meaningful policies, highlighting the continued difficulty of the task despite the increased amount of data.
Beyond success rate, RankQ also demonstrates a clear advantage in execution efficiency.
While CQL and Cal-QL reduce average task completion time from 12.5 seconds to 11.5 and 10.5 seconds, respectively, RankQ achieves a substantially faster execution time of just over 8 seconds.
This indicates that the benefits of improved value shaping extend beyond task success to more efficient and decisive policy behavior.

\bl{Sim-to-real evaluation.}
We evaluate sim-to-real transfer via zero-shot deployment of learned Cal-QL and RankQ cube stacking policies from Fig.~\ref{fig:vla-sim2real_results} on a real robot.
During online training, the fully trained RankQ policy learned a high-throughput behavior that dropped the cube from a height to minimize completion time.
While effective in the low-restitution simulator, this behavior transferred poorly to the real world due to the cube bouncing upon impact.
To avoid this sim-to-real mismatch, we also deploy an earlier checkpoint, RankQ (0.4), corresponding to 40\% of online training episodes from Fig.~\ref{fig:vla-sim2real_results}, prior to the emergence of this dropping behavior.
For the cube stacking experiments, we construct a 3$\times$3 grid, resulting in 72 possible cube configurations.
The baseline VLA policy, Cal-QL policy, and both RankQ checkpoints are evaluated for a total of 288 real-world experiments.
Results are summarized in Table~\ref{tab:sim2real_experiments}, where PSR denotes partial success rate (picking up the first cube), SR denotes overall success rate (stacking the cube without knocking it over after), and TTF denotes the average time-to-finish (with standard deviation) for successful rollouts.

Both Cal-QL and RankQ (0.4) achieve substantial improvements over the imitation baseline across all metrics, increasing overall success rate from 43.1\% to 84.7\% and 88.9\%, respectively.
\begin{wraptable}{r}{0.50\columnwidth} 
\vspace{-5pt} 
\renewcommand{\arraystretch}{1.1} 
\caption{Sim-to-real experiment results.} 
\centering 
\footnotesize 
\begin{tabular}{l | c c c} 
\toprule 
& PSR [$\uparrow$] & SR [$\uparrow$] & TTF (s) [$\downarrow$] \\ 
\midrule 
Baseline & $0.778$ & $0.431$ & \meanstd{14.81}{6.00}\\ 
Cal-QL & $\mathbf{1.000}$ & $0.847$ & \meanstd{14.75}{5.70}\\ 
RankQ (0.4) & $0.986$ & $\mathbf{0.889}$ & \meanstd{13.03}{5.85} \\ 
RankQ (1.0) & $0.986$ & $0.847$ & \meanstd{\mathbf{12.33}}{4.94} \\ 
\bottomrule 
\end{tabular} 
\label{tab:sim2real_experiments} 
\vspace{-5pt} 
\end{wraptable}
Partial success rate also improves markedly, increasing from 77.8\% to 100\% for Cal-QL and to 98.6\% for both RankQ checkpoints.
Despite corresponding to only 40\% of online training, RankQ (0.4) achieves near-perfect grasping performance comparable to Cal-QL while also attaining a slightly higher overall success rate.
In addition to improved success rates, the two RankQ checkpoints reduce average task completion time by 12.0\% and 16.7\%, respectively, whereas Cal-QL achieves similar speeds to the baseline policy.
Notably, despite the dropping behavior, the fully trained RankQ checkpoint still achieves a strong 84.7\% real-world success rate while attaining the fastest execution time.
Additional details and qualitative examples of sim-to-real deployment can be found in Sec.~\ref{sec:sim-to-real-examples}.

\section{Conclusion}
We introduced RankQ, an offline-to-online RL objective that augments temporal-difference learning with structured self-supervised action ranking.
Across D4RL benchmarks, RankQ achieved competitive overall performance, with particularly strong results on the challenging \texttt{antmaze-large} environments.
We further demonstrated strong performance for offline-to-online VLA RL fine-tuning in both low-data and high-data regimes, where RankQ consistently achieved the highest final success rates and, in the high-data setting, the fastest task completion time.
Finally, we showed successful sim-to-real transfer for cube stacking, improving real-world success rate from 43.1\% to 88.9\%.
Overall, our results suggest that explicitly modeling relative action ordering provides an effective alternative to uniformly pessimistic value estimation.

\bl{Limitations and future work.}
Despite the strong empirical results, RankQ has several limitations.
First, while most hyperparameters were kept constant across environments to demonstrate generality, the optimal configuration may still require environment-specific tuning.
Second, RankQ depends on sparse success signals, limiting online training to completed trajectories.
Finally, while RankQ appears to improve the Q-landscape, this work does not provide formal theoretical guarantees relating action ranking consistency to policy improvement and also assumes that action-space proximity to successful behavior is, on average, correlated with action quality.
This heuristic may be violated in highly multimodal or discontinuous action spaces, where nearby actions can correspond to qualitatively different behaviors.
However, this concern is partially mitigated in our setting by Gaussian policy optimization, which typically concentrates probability mass around a single mode rather than simultaneously optimizing across widely separated action modes.
For future work, we plan to study RankQ in even larger action spaces (e.g., bimanual manipulation), as well as its real-world RL sample efficiency.

\newpage

\bibliography{references}  

\newpage

\appendix
\counterwithin{table}{section}
\counterwithin{figure}{section}
\counterwithin{equation}{section}
\renewcommand{\thetable}{\thesection.\arabic{table}}
\renewcommand{\thefigure}{\thesection.\arabic{figure}}
\renewcommand{\theequation}{\thesection.\arabic{equation}}

\section{RankQ Loss}
\label{sec:rankq_loss}
To formulate the RankQ loss, we use the softplus function $\textrm{sp}(x) = \log(1 + e^x)$, which provides a smooth approximation to the hinge loss.
We define the pairwise ranking function
\begin{equation}
    \mathcal R(s, a^+, a^-) = \textrm{sp}(Q_\theta(s, a^-) - Q_\theta(s, a^+)).
\end{equation}
We first formulate the ordering constraints from Eq.~\ref{eq:success_ranks}:
\[
Q(s,a) > Q(s,a'), \quad
a' \in \{a_{\epsilon}, a_{2\epsilon}, a_r, a_p\}, \;
(s,a) \sim \mathcal{D}_{\mathrm{success}}
\]
as the loss
\begin{equation}
\begin{aligned}
    \mathcal L_Q^\textrm{succ}(\theta) =
    \mathbb E_{s,a \sim \mathcal D_\textrm{success}} \Bigg[
    &\mathbb E_{\epsilon \sim \mathcal N(0, \sigma^2)} \Big[
        \mathcal R(s, a, a+\epsilon)
        + \mathcal R(s, a, a+2\epsilon)
    \Big] + \\
    &\mathbb E_{a_r \sim \mathcal U(-1, 1)^{|a|}} \Big[
        \mathcal R(s, a, a_r)
    \Big] + \\
    &\mathbb E_{a_p \sim \mathcal D} \Big[
        \mathcal R(s, a, a_p)
    \Big]
    \Bigg],
\end{aligned}
\label{eq:succ_loss}
\end{equation}
where $\sigma$ determines the standard deviation of perturbations.
Similarly, we formulate the chained ordering constraints from Eq.~\ref{eq:success_neg_ranks}:
\[
Q(s,a_{\epsilon}) > Q(s,a_{2\epsilon}) > Q(s,a_r)
\]
as the loss
\begin{equation}
\begin{aligned}
    \mathcal L_Q^\textrm{chain}(\theta) =
    \mathbb E_{s,a \sim \mathcal D_\textrm{success}} \Bigg[
    &\mathbb E_{\epsilon \sim \mathcal N(0, \sigma^2), \,
    a_r \sim \mathcal U(-1, 1)^{|a|}} \Big[
        \mathcal R(s, a+\epsilon, a+2\epsilon) + \mathcal R(s, a+2\epsilon, a_r)
    \Big]
    \Bigg].
\end{aligned}
\label{eq:neg_loss}
\end{equation}

Finally, the failure-based ordering constraints from Eq.~\ref{eq:failure_ranks}:
\[
Q(s,a) > Q(s,a_r), \quad
(s,a)\sim \mathcal D_\textrm{failure}
\]
are formulated as
\begin{equation}
\begin{aligned}
    \mathcal L_Q^\textrm{fail}(\theta) =
    \mathbb E_{s,a \sim \mathcal D_\textrm{failure}} \Bigg[
    &\mathbb E_{a_r \sim \mathcal U(-1, 1)^{|a|}} \Big[
        \mathcal R(s, a, a_r)
    \Big]
    \Bigg].
\end{aligned}
\label{eq:fail_loss}
\end{equation}

Combining Eqs.~\ref{eq:succ_loss}, \ref{eq:neg_loss}, and~\ref{eq:fail_loss}, we obtain the final RankQ objective:
\begin{equation}
\label{eq:rank_q_loss}
    \mathcal L_Q^\textrm{RankQ}(\theta)
    =
    \green{
    \alpha_0
    \left(
        \mathcal L_Q^\textrm{succ}(\theta)
        +
        \mathcal L_Q^\textrm{chain}(\theta)
    \right)
    +
    \alpha_1
    \mathcal L_Q^\textrm{fail}(\theta) } + \mathcal L_\textrm{TD}(\theta),
\end{equation}
where $\alpha_0$ and $\alpha_1$ are hyperparameters controlling the influence of success and failure trajectories, respectively.
In this work, we use default values of $\alpha_0 = \alpha_1 = 1$ and $\sigma = 0.15$ for most experiments.

\bl{Toy example formulation.}
To construct the toy example in Fig.~\ref{fig:toy_example}, we formulate a regression problem with a fixed state and a two-dimensional continuous action space $a = (a_0, a_1) \in [-1, 1]^2$.
The environment contains a circular success region centered at the origin defined by $\lVert a \rVert \leq R$, where $R$ denotes the success radius.
Offline success actions are uniformly sampled within the circle and assigned a reward of 1, while failure actions are sampled outside the circle and to the right of the success region with a reward of 0.
A simple MLP critic network is then trained on this offline dataset using each critic loss formulation.

\newpage

\bl{Pseudocode.}
Python pseudocode for computing Eq.~\ref{eq:rank_q_loss}.

\begin{minted}[
    frame=single,
    framesep=2mm,
    fontsize=\footnotesize,
    breaklines=true
]{python}
import torch
import torch.nn.functional as F


def compute_rankq_loss(critic: torch.nn.Module, state: torch.Tensor,
        rollout_action: torch.Tensor, success_mask: torch.Tensor,
        noise_scale: float = 0.15, alpha_0: float = 1.0, alpha_1: float = 1.0):

    bs = rollout_action.shape[0]
    success_indices = torch.where(success_mask)[0]
    failure_indices = torch.where(~success_mask)[0]
    eps = torch.randn_like(rollout_action)
    perm_indices = torch.roll(torch.arange(bs), shifts=-1)

    # Define all types of actions
    all_actions = {
        "rollout": rollout_action,
        "noisy": rollout_action + eps * noise_scale,
        "very_noisy": rollout_action + eps * 2 * noise_scale,
        "permuted": rollout_action[perm_indices],
        "random": torch.rand_like(rollout_action) * 2 - 1,
    }

    # Compute Q-values for all actions
    q_values = {
        name: critic(state, action)
        for name, action in all_actions.items()
    }

    pairs = [
        # L^fail
        ("rollout", "random"),
        # L^succ
        ("rollout", "noisy"), ("rollout", "very_noisy"),
        ("rollout", "permuted"), ("rollout", "random"),
        # L^chain
        ("noisy", "very_noisy"), ("very_noisy", "random")
    ]

    rankq_loss = torch.zeros(bs)
    for i, (pos_type, neg_type) in enumerate(pairs):
        ind = failure_indices if i == 0 else success_indices
        alpha = alpha_1 if i == 0 else alpha_0
        
        q_pos = q_values[pos_type][ind]
        q_neg = q_values[neg_type][ind]

        rankq_loss[ind] += alpha * F.softplus(q_neg - q_pos)

    return rankq_loss.mean()
\end{minted}

\newpage

\section{Training Setup and Parameters}
\label{sec:training_setup}
In this section, we list the relevant training hyperparameters for all environments.
All D4RL experiments are run on a single NVIDIA RTX 4090 while VLA experiments are run on 8 NVIDIA RTX 6000 Ada GPUs.
For \texttt{vla-sim2real-cube-stacking}, RankQ and SAC+OFF take about 12 hours to train, while CQL and Cal-QL take about 20 hours (details on the reason for this discrepancy are provided below).

\bl{Reproducibility disclaimer.}
When running the D4RL training code released by~\citet{nakamoto2023calql} directly without modification, we were unable to replicate the performance of CQL and Cal-QL originally reported in the Cal-QL paper.
Contrary to the paper, we found that CQL and Cal-QL exhibited very high variance across random seeds in several D4RL environments.
We also observed that several baseline algorithms (e.g., SAC+OFF and Hybrid RL), which were originally reported to achieve near-zero success on certain environments, were able to attain substantially higher performance---and in some cases near-maximal success---with sufficient training.
For example, in our experiments, Hybrid RL and SAC+OFF achieve near-100\% success on \texttt{antmaze-medium-diverse}, whereas the original paper reports near-0\% and roughly 65\% success, respectively.
Finally, though D4RL contains additional environments such as \texttt{franka-kitchen}, these were omitted because the source code from~\citet{nakamoto2023calql} did not include them.

\bl{Common hyperparameters.}
During tuning, we observed that several algorithms exhibited instability and occasionally diverged during training on the \texttt{adroit-door} and \texttt{adroit-relocate} environments.
For these two environments, we introduced gradient norm clipping, increased the mini-batch size, and lowered the learning rates to improve stability.
With these modifications, no algorithms experienced performance collapse during training, and several achieved substantially higher performance (Fig.~\ref{fig:d4rl_results}) than originally reported in~\citep{nakamoto2023calql}.
Table~\ref{tab:common_hyperparameters} summarizes the hyperparameters shared across algorithms for each environment.
\begin{table}[h]
\renewcommand{\arraystretch}{1.1}
\caption{Common hyperparameters.}
\centering
\footnotesize
\begin{tabular}{l|c c c c c}
\toprule
Environment & mini-batch size & actor lr & critic lr & grad norm clip & replay buffer length \\
\midrule
\texttt{antmaze-all} 
& 256 
& 1e-4
& 3e-4
& 1.0
& 1,000,000 \\

\texttt{adroit-pen} 
& 256 
& 1e-4
& 3e-4
& 1.0 
& 1,000,000 \\

\texttt{adroit-door} 
& 512 
& 2e-5
& 1e-4
& 1.0 
& 1,000,000 \\

\texttt{adroit-relocate} 
& 512 
& 2e-5
& 1e-4
& 1.0 
& 1,000,000 \\

\texttt{vla-low-data-all}
& 960 
& 2e-5
& 1e-4
& 0.5 
& 500,000 \\

\texttt{vla-sim2real} 
& 1920 
& 2e-5
& 1e-4
& 0.5 
& 500,000 \\

\bottomrule
\end{tabular}
\label{tab:common_hyperparameters}
\end{table}

A list of all tuned CQL, Cal-QL, and RankQ-specific parameters can be seen in Tables~\ref{tab:cql_hyperparameters} and~\ref{tab:rankq_hyperparameters}, respectively.
Given that RankQ operates through relative action comparisons rather than absolute value calibration, it requires minimal tuning to work effectively.

\begin{figure}[h]
\includegraphics[width=\columnwidth]{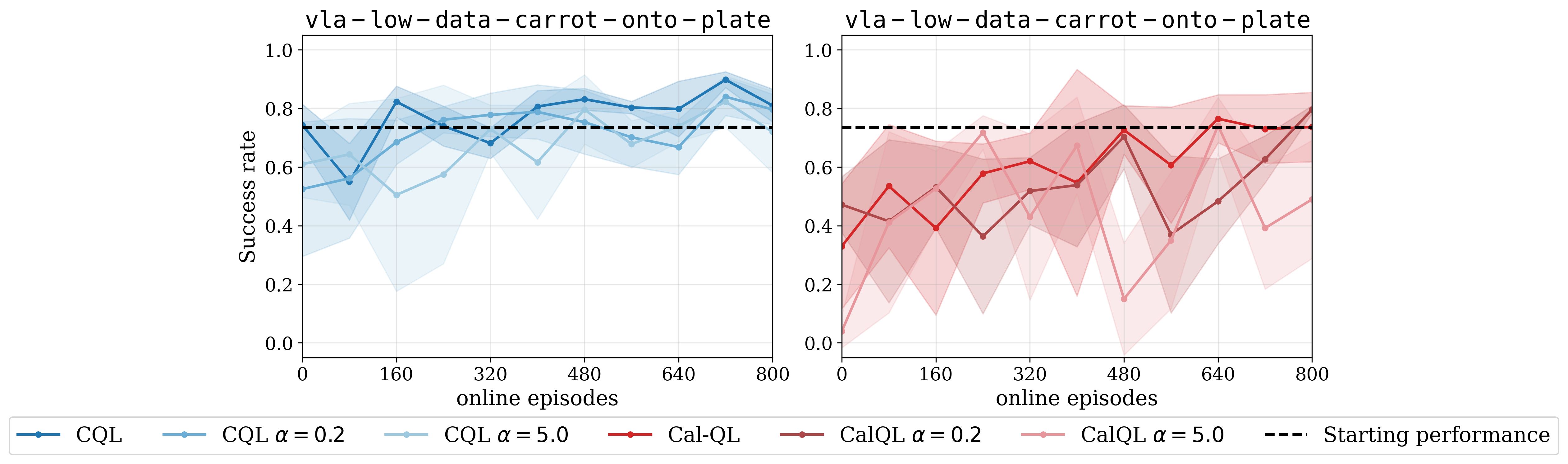}
\caption{
    CQL and Cal-QL $\alpha$ parameter tuning using \texttt{vla-low-data-carrot-onto-plate}.
}
\label{fig:cql_calql_tune}
\end{figure}

\bl{Regularization weights.}
For CQL and Cal-QL, we use the same $\alpha$ values as~\citep{nakamoto2023calql} for the D4RL environments.
For the VLA environments, we use \texttt{vla-low-data-carrot-onto-plate} to tune the $\alpha$ for both CQL and Cal-QL (Fig.~\ref{fig:cql_calql_tune}).
As this parameter had minimal impact on performance, we ultimately use $\alpha = 1.0$ for both methods.
For RankQ, we set $\alpha_1 = 1$ across all benchmarks.
For $\alpha_0$, we use a value of 20 for the \texttt{antmaze} environments and 1 otherwise.
The choice of $\alpha_0 = 20$ is motivated by the dataset structure of the \texttt{antmaze} environments.
Although the majority of trajectories are successful, failure trajectories terminate via timeout after 1000 steps, resulting in successful transitions comprising less than 5\% of samples in a typical offline mini-batch.
This motivates using $\alpha_0 = 20$ to prevent the influence of failure transitions from overwhelming the successful ones.

\bl{Mixing ratio.}
The mixing ratio denotes the proportion of offline samples relative to online samples within each mini-batch.
For example, a mixing ratio of 0.75 indicates that 75\% of a mini-batch is sampled from the offline dataset and 25\% from the online replay buffer.
A mixing ratio of $-1$ denotes a setup in which online samples are simply appended to the offline dataset, causing the influence of offline data to naturally diminish over the course of online training.
For CQL and Cal-QL, we use the tuned values from~\citep{nakamoto2023calql} for the D4RL environments.
For RankQ on D4RL, we perform a similar tuning over $\{-1, 0.5\}$ for the \texttt{adroit} environments.
Table~\ref{tab:rankq_hyperparameters} shows that RankQ actually performs better with a mixing ratio of $-1$ on certain tasks, suggesting that it is less reliant on offline data during online training.
For the VLA environments, since the dataset consists solely of the policy's own rollouts, we use a mixing ratio of $-1$ across all algorithms.

\begin{table}[h]
\renewcommand{\arraystretch}{1.1}
\caption{CQL and Cal-QL hyperparameters. 
Metrics wrapped in brackets showcase the swept parameters with the best performing one being bolded. 
Metrics denoted with $*$ are hyperparameters taken directly from the tuning of~\citep{nakamoto2023calql}.}
\centering
\footnotesize
\begin{tabular}{l|c c c }
\toprule
Environment & $\alpha$ & target action gap & mixing ratio  \\
\midrule
\texttt{antmaze-all} 
& - 
& 0.8 
& 0.5 \\

\texttt{adroit-all} 
& 1 
& - 
& \{ -1, 0.25, \textbf{0.5} \}*  \\

\texttt{vla-carrot-on-plate} 
& \{ 0.2, \textbf{1}, 5 \} 
& - 
& -1 \\

\texttt{vla-cube-stack} 
& 1
& - 
& -1 \\

\texttt{vla-spoon-in-bowl} 
& 1 
& - 
& -1 \\

\bottomrule
\end{tabular}
\label{tab:cql_hyperparameters}
\end{table}

\begin{table}[h]
\renewcommand{\arraystretch}{1.1}
\caption{RankQ hyperparameters.
Metrics wrapped in brackets showcase the swept parameters with the best performing one being bolded. 
}
\centering
\footnotesize
\begin{tabular}{l|c c c}
\toprule
Environment & $(\alpha_0, \alpha_1)$ & $\sigma$ & mixing ratio \\
\midrule
\texttt{antmaze-all} 
& \{ (1, 1), \textbf{(20, 1)} \} 
& 0.15 
& 0.5 \\

\texttt{adroit-pen} 
& (1, 1) 
& 0.15 
& \{ \textbf{-1}, 0.5 \} \\

\texttt{adroit-door} 
& (1, 1)
& 0.15 
& \{ \textbf{-1}, 0.5 \} \\

\texttt{adroit-relocate} 
& (1, 1)
& 0.15 
& \{ -1, \textbf{0.5} \} \\

\texttt{vla-low-data-all} 
& (1, 1) 
& 0.15 
& -1 \\

\texttt{vla-sim2real} 
& (1, 1) 
& 0.15 
& -1 \\

\bottomrule
\end{tabular}
\label{tab:rankq_hyperparameters}
\end{table}

\bl{Number of critic calls.}
Given the $\mathbb E_{s \sim \mathcal D, a \sim \pi(\cdot \mid s)}$ expectation term in the CQL (Eq.~\ref{eq:cql}) and Cal-QL (Eq.~\ref{eq:calql}) regularizers, multiple critic evaluations are required to accurately estimate the expectation.
For the \texttt{vla} environments, we use 10 policy-sampled actions and 10 randomly sampled actions to compute this expectation, following the setup used for the D4RL environments in~\citep{nakamoto2023calql}, resulting in a total of 20 batched critic evaluations per training iteration.
In comparison, RankQ requires only 4 batched critic evaluations, corresponding to the four suboptimal actions defined in Eq.~\ref{eq:success_ranks}.

For shallow MLPs, this additional computation does not noticeably impact training speed.
However, for larger VLA models, it introduces significant computational overhead.
In our experiments, the 20 batched critic evaluations required us to halve the mini-batch size and use gradient accumulation due to increased GPU memory usage.
Combined with the $5\times$ larger number of critic evaluations, this caused the backward pass of CQL and Cal-QL to be approximately $2.8\times$ slower than RankQ on the \texttt{vla} environments.
On the \texttt{vla-sim2real-cube-stacking} environment, this resulted in an overall training wall-clock time increase of approximately 60\%.

\bl{VLA domain randomization.}
For the \texttt{vla-sim2real-cube-stacking} environment, we use the same domain randomization parameters and 100 EmbodiedGen~\citep{wang2025embodiedgengenerative3dworld, wang2026embodiedgenv2agenticsimulationready} scenes as~\citep{choi2026scalingsimtorealreinforcementlearning}.
The parameter values are provided in Table~\ref{tab:domain_randomization}.
For the \texttt{vla-low-data} environments, we use a single EmbodiedGen scene together with the same object xy-position and yaw randomizations listed in Table~\ref{tab:domain_randomization}.
\begin{table}[h]
\renewcommand{\arraystretch}{1.1}
\caption{Domain randomization ranges for \texttt{vla} environments.}
\centering
\footnotesize
\begin{tabular}{ll|ll}
\toprule
Parameter & Value & Parameter & Value \\
\midrule
object x-position (m) & $[0.2, 0.4]$ & robot z-height perturb (m) & $[0, 0.05]$ \\
object y-position (m) & $[-0.15, 0.15]$ & robot joint pos perturb (rad) & $[-0.1, 0.1]^6$ \\
object yaw orientation (rad) & $[0, 2\pi]$ & camera xyz-position (m) & $[-0.05, 0.05]^3$ \\
ambient light RGB color & $[0, 0.6]^3$ & directional light brightness & $[0.5, 1.5]$ \\
\bottomrule
\end{tabular}
\label{tab:domain_randomization}
\end{table}

\newpage

\section{SACFlow Formulation}
\label{sec:sac_flow}

For the VLA, we use action chunks $\mathbf A \in \mathbb R^{C\times 7}$ as the MDP action, where $C$ denotes the action chunk size.
Observations $\mathbf s_t$ consist of an RGB image from a static external camera together with a language instruction.
For all VLA experiments, we use an action chunk size of $C=4$ with 5\,Hz control.

\bl{Flow matching models.}
Similar to the setup in~\citep{choi2026scalingsimtorealreinforcementlearning}, we use $\pi_0$~\citep{black2024pi0visionlanguageactionflowmodel}, which consists of a VLM backbone $E_\theta$ and a flow-matching action head $v_\theta$.
The model is trained using the rectified flow-matching objective
\begin{equation}
    \mathcal L_\textrm{flow}(\theta) =
    \mathbb E_{\mathbf s_t,\mathbf A_t^1\sim\mathcal D, \boldsymbol \epsilon \sim \mathcal N(\mathbf 0, \mathbf I), \tau \sim \mathcal U(0, 1)}
    \left[
    \left\lVert
    v_\theta(\mathbf A_t^\tau, KV_\theta(\mathbf s_t), \tau)
    -
    (\mathbf A_t^1 - \boldsymbol \epsilon)
    \right\rVert_2^2
    \right],
\end{equation}
where $\mathcal D$ is the demonstration dataset, $\tau \in [0,1]$ denotes the continuous integration time, $KV_\theta(\mathbf s_t)$ are the key-value tensors of $E_\theta(\mathbf s_t)$, and $\mathbf A_t^\tau$ is defined as
\begin{equation}
    \mathbf A_t^\tau = \tau \mathbf A_t^1 + (1-\tau)\boldsymbol \epsilon.
\end{equation}
Actions are generated by numerically integrating the learned ODE
\begin{equation}
    \frac{d\mathbf A_t^\tau}{d\tau}
    =
    v_\theta(\mathbf A_t^\tau, KV_\theta(\mathbf s_t), \tau).
\end{equation}

\bl{SACFlow.}
To make flow-matching models amenable to SAC-style updates, learnable noise $\sigma_\phi(\mathbf A_t^\tau, \mathbf z_t, \tau)$ is injected into the numerical integration process to produce a stochastic policy, where $\mathbf z_t$ is the hidden state of the VLM backbone defined as $\mathbf z_t = E_\theta(\mathbf s_t)$.
Denoting the number of integration steps as $K = 1 / \Delta \tau$, each integration step becomes a Gaussian sample of the form
\begin{align}
    &\hat{\mathbf A}
    =
    \mathbf A_t^\tau
    +
    v_\theta(\mathbf A_t^\tau, KV_\theta(\mathbf s_t), \tau)\Delta\tau, \\
    &\mathbf A_t^{\tau + \Delta\tau}
    \sim
    \mathcal N(\hat{\mathbf A},
    \sigma_\phi(\mathbf A_t^\tau, \mathbf z_t, \tau)),
\end{align}
which can be chained together to compute the joint log-probability of the full denoising trajectory for entropy regularization.
For the critic, we use an MLP Q-value head $Q_\psi(\mathbf z_t, \mathbf A)$ to estimate action values.
For all VLA experiments, we omit entropy regularization from both the critic and actor losses, and instead constrain $\log \sigma_\phi$ to the range $[-2.5, -2.0]$.
Furthermore, we RL fine-tune the VLA using only a single integration step ($K=1$), effectively reducing inference latency by more than $2\times$~\citep{choi2026scalingsimtorealreinforcementlearning}.
Finally, we fine-tune only the action head $v_\theta$ while freezing the VLM backbone $E_\theta$.

\newpage

\section{D4RL RankQ Ablation}
\label{sec:rankq_ablation}
\begin{figure}[h]
\includegraphics[width=\columnwidth]{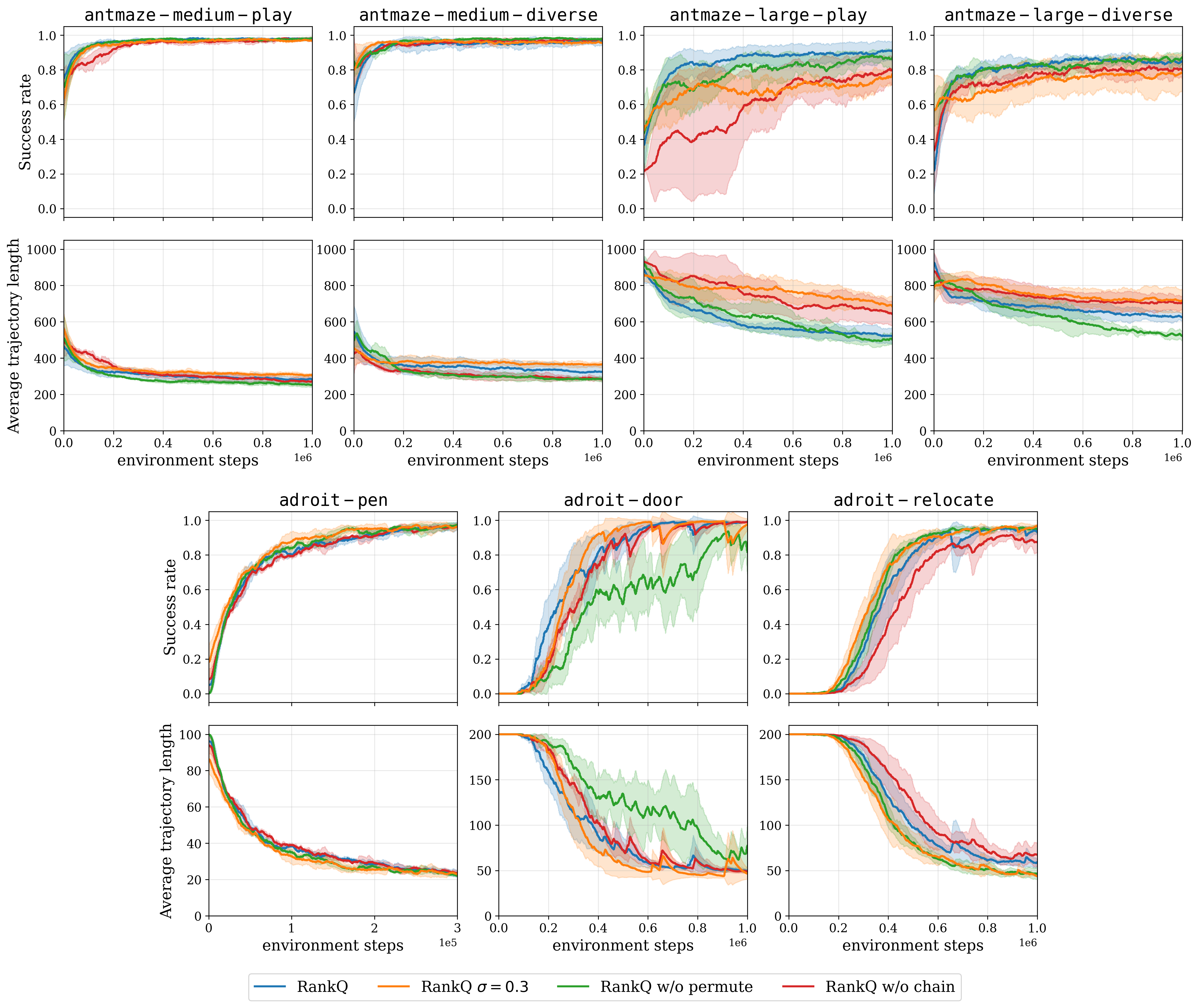}
\caption{
RankQ ablation results on D4RL environments.
}
\label{fig:d4rl_rankq_ablation}
\end{figure}

In this section, we ablate key components of the RankQ loss (Sec.~\ref{sec:rankq_loss}) on D4RL.
To do so, we rerun the D4RL benchmarks and evaluate the following ablations:
\begin{enumerate}[leftmargin=12pt, topsep=0pt, itemsep=0pt]
    \item Increasing the noisy perturbation scale $\sigma$ from 0.15 to 0.30.
    \item Omitting permuted-action ranking $a_p$ from $\mathcal L_Q^\textrm{succ}(\theta)$.
    \item Omitting the chain loss $\mathcal L_Q^\textrm{chain}(\theta)$.
\end{enumerate}

As shown in Fig.~\ref{fig:d4rl_rankq_ablation}, the effect of each ablation varies across environments.
Most notably, the easiest environments (\texttt{antmaze-medium} and \texttt{adroit-pen}) exhibit only minor performance differences between ablations.
This changes as the difficulty of the environments increases.
For \texttt{antmaze-large-play}, the original RankQ formulation achieves the strongest and most stable performance, while increasing $\sigma$ and omitting $\mathcal L_Q^\textrm{chain}(\theta)$ both produce noticeable performance degradation.
For \texttt{antmaze-large-diverse}, omitting permuted-action ranking results in a slightly more efficient policy compared to the full formulation, although success rates remain comparable.
In contrast, omitting permuted-action ranking produces a noticeable negative effect on \texttt{adroit-door}, while the remaining ablations perform comparably to the base formulation.
Finally, for \texttt{adroit-relocate}, omitting $\mathcal L_Q^\textrm{chain}(\theta)$ negatively impacts performance, whereas increasing $\sigma$ and omitting permuted-action ranking produce slightly more efficient policies than the baseline, though success rates again remain comparable.
Overall, these experiments suggest that the optimal RankQ configuration may vary depending on the environment, although the default formulation consistently provides strong baseline performance.

\newpage

\section{Q-landscape Analysis}
\label{sec:q-landscape-analysis}

In this section, we analyze various statistics of the Q-values learned under different critic formulations for the \texttt{vla-sim2real-cube-stacking} experiments shown in Fig.~\ref{fig:vla-sim2real_results}.

\begin{figure}[h]
\centering
\includegraphics[width=\columnwidth]{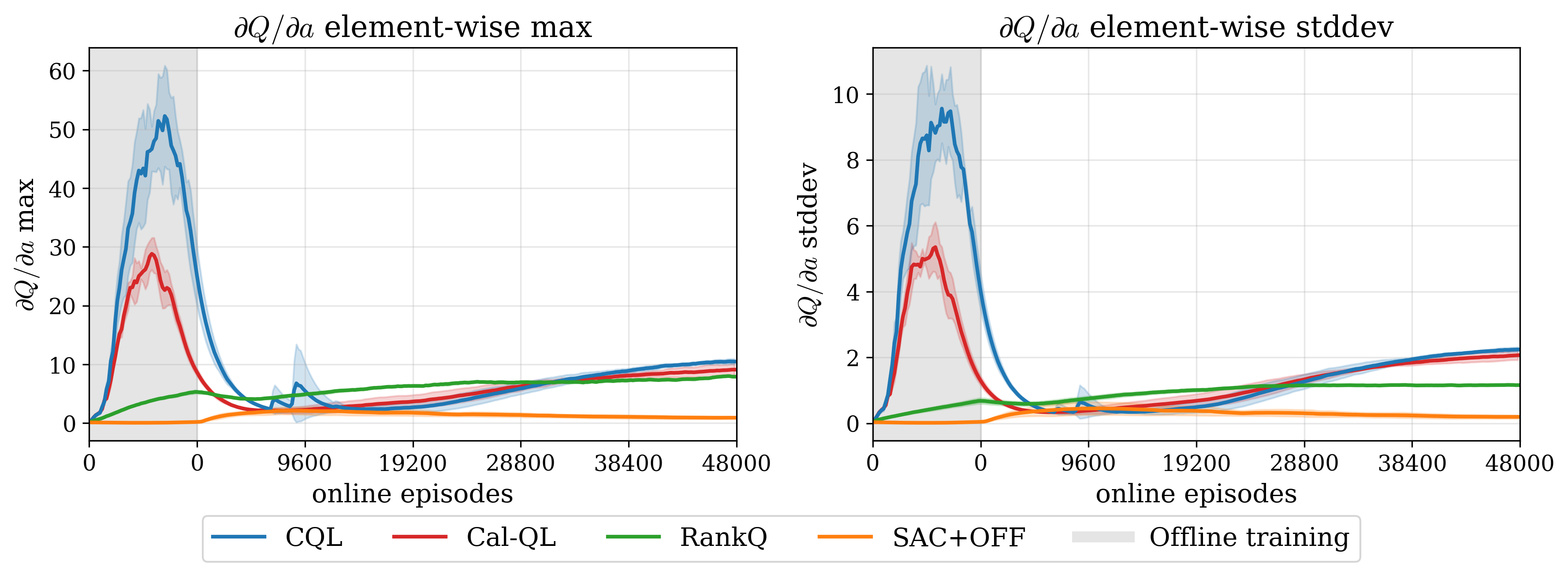}
\caption{
The $\partial Q / \partial a$ statistics for the \texttt{vla-sim2real-cube-stacking} experiments from Fig.~\ref{fig:vla-sim2real_results}.
The gray shaded region denotes the offline RL phase.
}
\label{fig:dqda_results}
\end{figure}

To study the influence of different critic formulations on $\partial Q / \partial a$, we plot the element-wise maximum and standard deviation of $\partial Q / \partial a$ throughout offline-to-online training in Fig.~\ref{fig:dqda_results}.
We observe that methods such as CQL and Cal-QL produce substantially sharper gradient landscapes during the offline RL phase due to their stronger pessimism.
CQL exhibits the largest spikes as a result of its unbounded pessimism, while Cal-QL produces smaller spikes due to its calibration floor.

Although these spikes eventually decrease during offline training, transitioning to online fine-tuning near their peak could negatively impact optimization stability.
In contrast, RankQ produces substantially more stable statistics and exhibits minimal distributional shift when transitioning from offline to online training.
SAC+OFF fails to achieve meaningful success rates (Fig.~\ref{fig:vla-sim2real_results}), making its $\partial Q / \partial a$ statistics comparatively uninformative.

\newpage

\section{Critic Accuracies}
\label{sec:critic-accuracies}

\begin{figure}[h]
\centering
\includegraphics[width=\columnwidth]{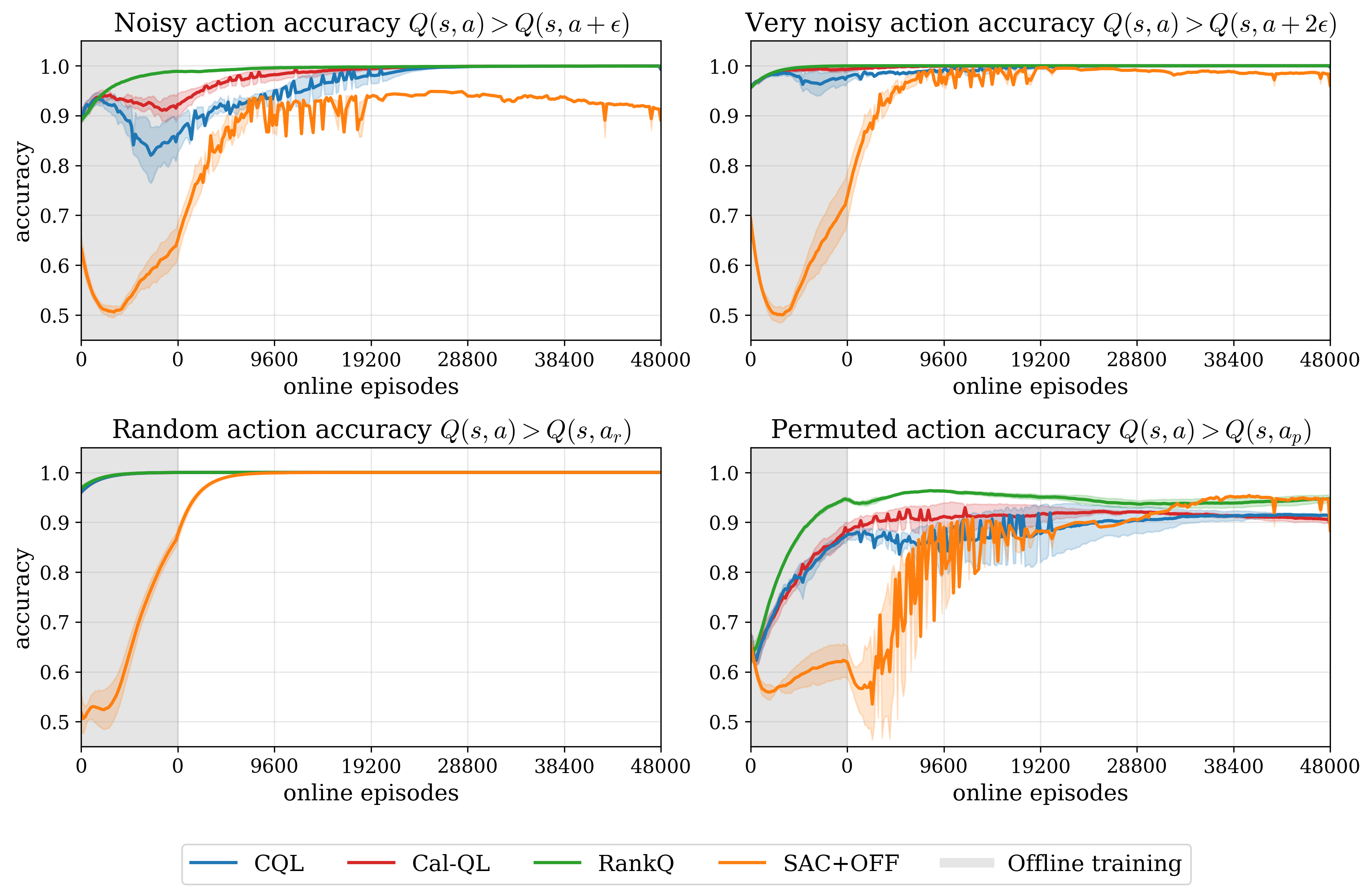}
\caption{
Accuracy results comparing Q-values of successful actions against the four categories of suboptimal actions defined in Eq.~\ref{eq:success_ranks} for the \texttt{vla-sim2real-cube-stacking} experiments from Fig.~\ref{fig:vla-sim2real_results}.
}
\label{fig:q_acc_results}
\end{figure}

We also study the influence of different critic formulations on the accuracy of ranking successful actions above the four categories of suboptimal actions shown in Fig.~\ref{fig:q_acc_results}.
As expected, RankQ achieves the highest accuracies across all four categories due to its explicit ordering constraints.

CQL and Cal-QL achieve comparably high accuracies for random and very noisy actions, but exhibit lower accuracies than RankQ for noisy and permuted actions.
As training progresses, CQL and Cal-QL eventually approach RankQ's noisy-action accuracy.
SAC+OFF exhibits the lowest and slowest-improving accuracies since it relies solely on TD learning.

Overall, these results suggest that the suboptimal action categories defined by RankQ capture meaningful structure in the action space.
Notably, other offline RL methods such as CQL and Cal-QL also implicitly learn similar action rankings, albeit at a substantially slower rate.
This further supports the idea that RankQ's simple action-ordering objective is aligned with the implicit behavior of existing state-of-the-art methods while achieving higher sample efficiency.

\newpage

\section{Visualization of VLA Training Environments}
\begin{figure}[h]
\centering
\includegraphics[width=0.9\columnwidth]{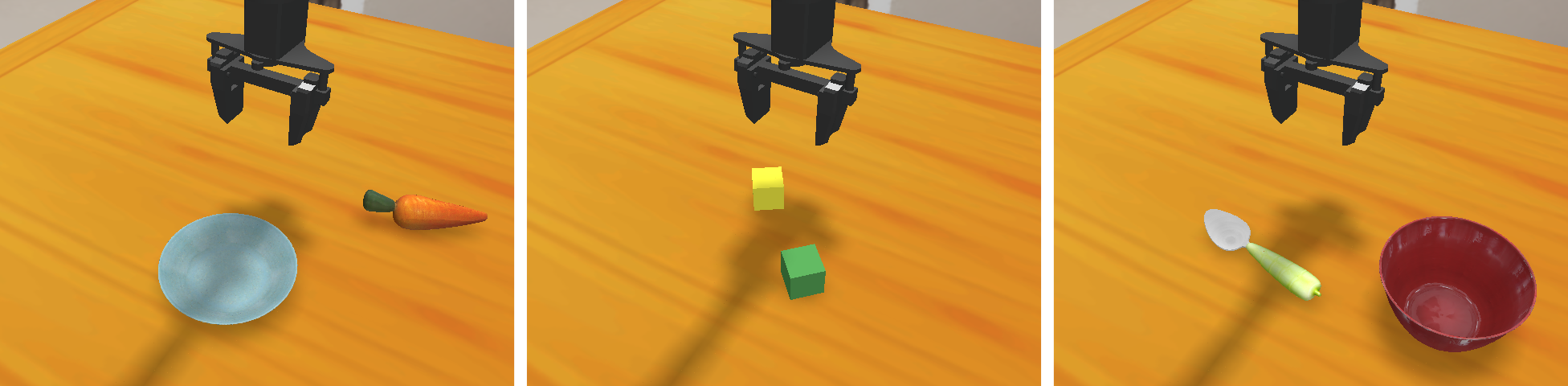}
\caption{
Visualization of the \texttt{vla-low-data} environments.
}
\label{fig:low_data_scenes}
\end{figure}

\begin{figure}[h]
\centering
\includegraphics[width=\columnwidth]{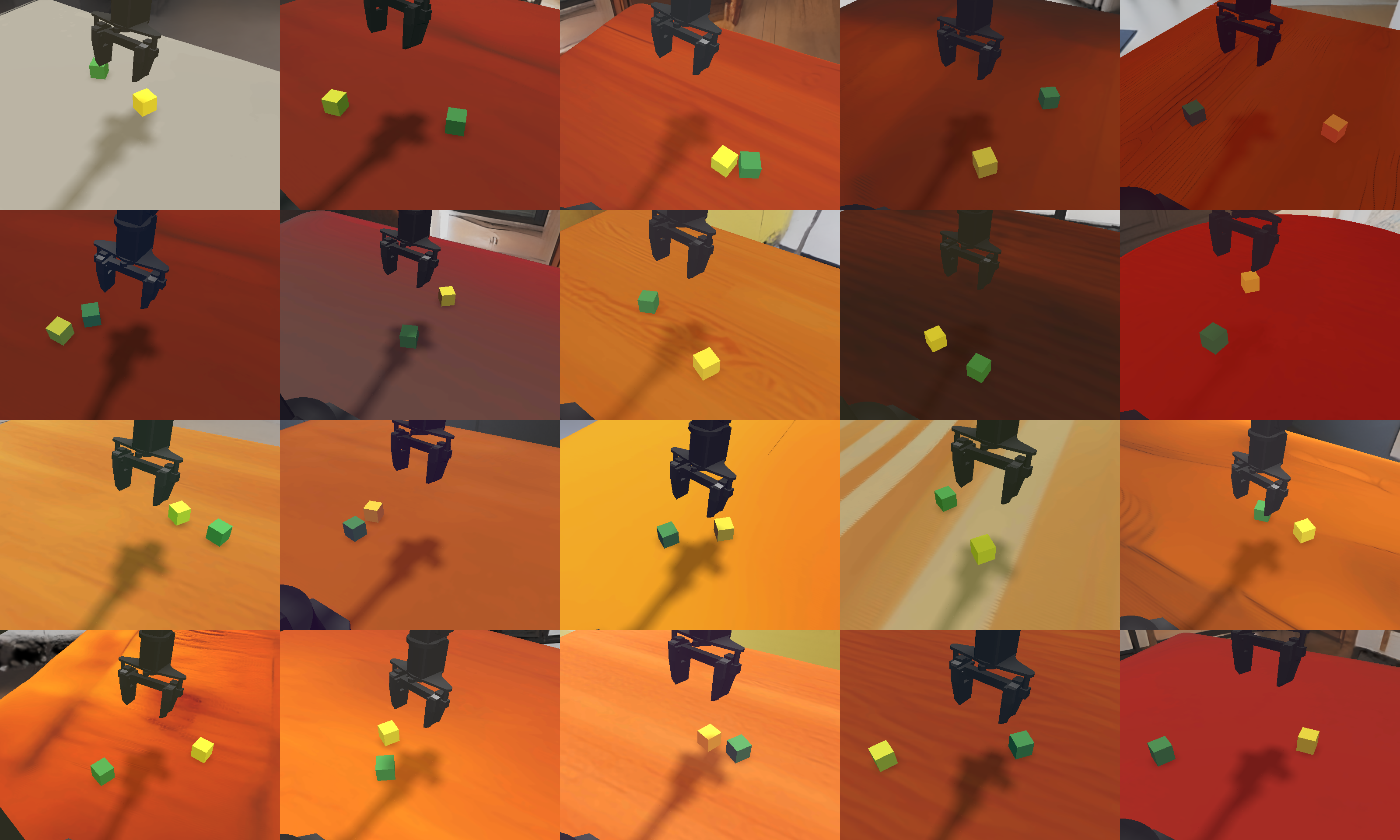}
\caption{
Visualization of the \texttt{vla-sim2real-cube-stacking} environment. 
Each frame shows the start of an episode where scene, robot pose, camera pose, and lighting randomizations can clearly be seen.
}
\label{fig:high_data_scenes}
\end{figure}

In this section, we show qualitative examples of the \texttt{vla-low-data} and \texttt{vla-sim2real} environments in Figs.~\ref{fig:low_data_scenes} and \ref{fig:high_data_scenes}, respectively.
For all environments, we use the ManiSkill 3 simulation framework~\citep{taomaniskill3}.

\newpage

\section{Sim-to-Real Settings and Examples}
\label{sec:sim-to-real-examples}

\begin{figure}[h]
\centering
\includegraphics[width=0.81\columnwidth]{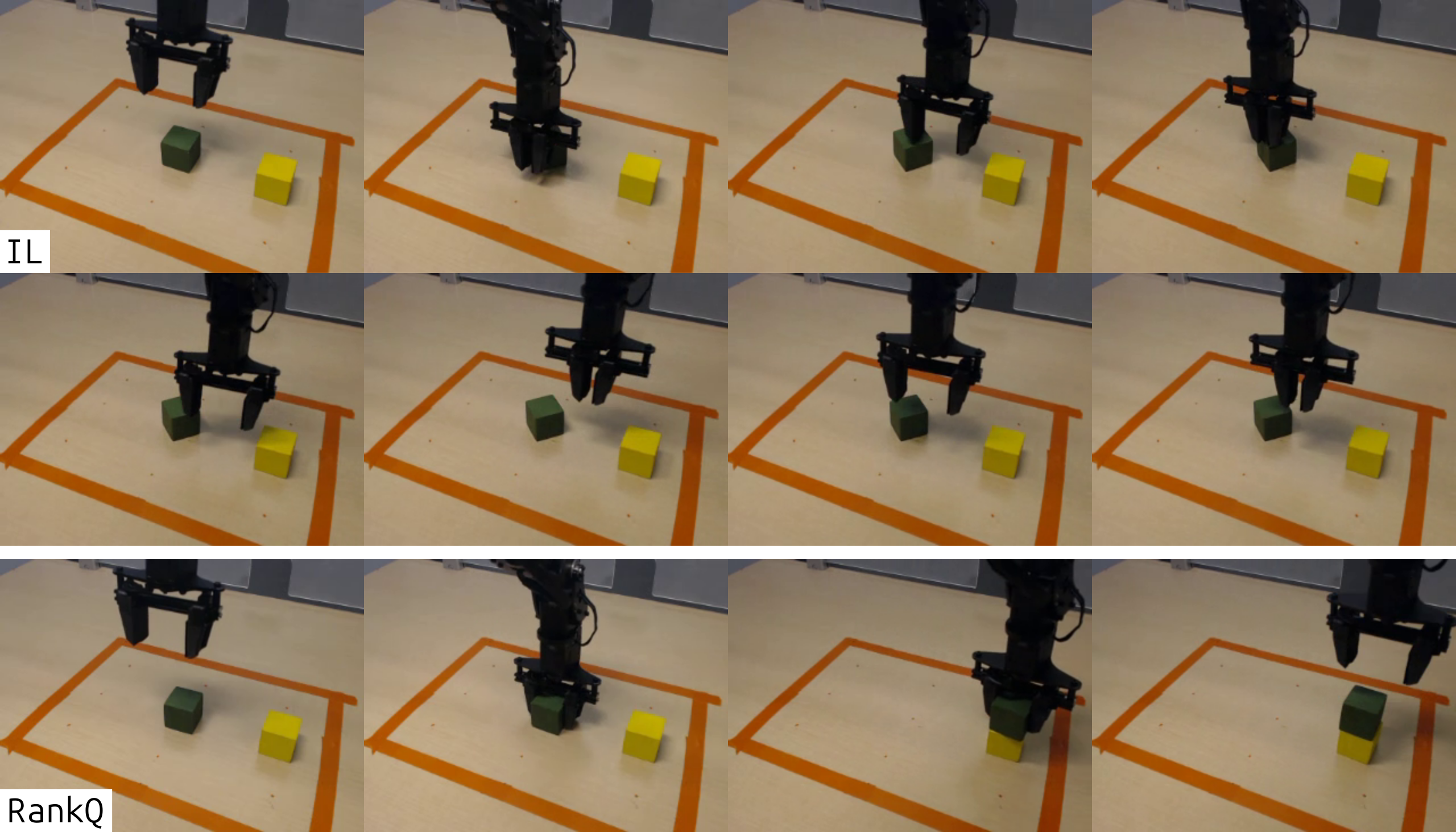}
\caption{
Rollout example where the baseline policy repeatedly fails to grasp the green cube until timeout.
In comparison, RankQ (0.4) stacks the cubes in one attempt.
}
\label{fig:sim2real_snapshot_0}
\end{figure}

\begin{figure}[h]
\centering
\includegraphics[width=0.81\columnwidth]{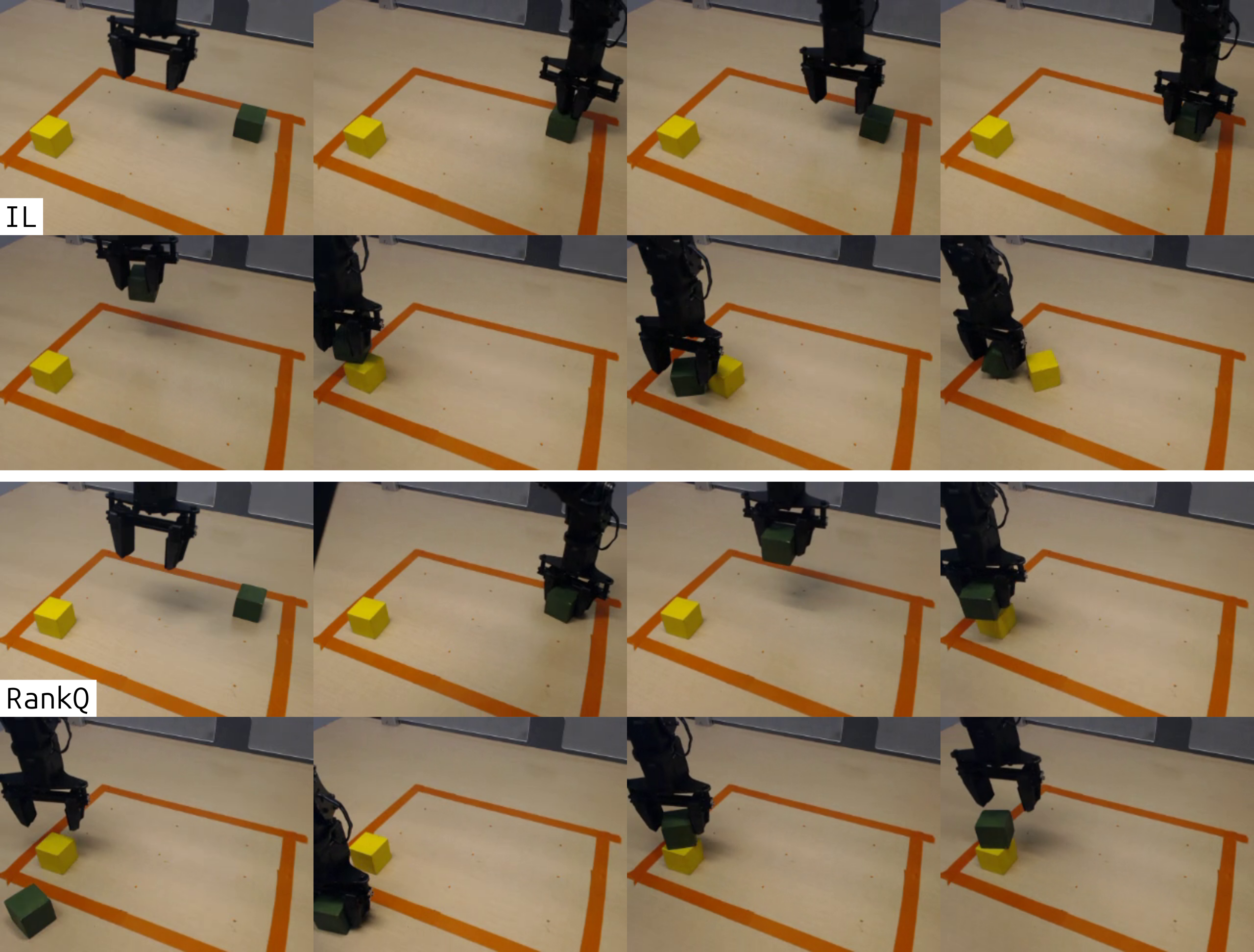}
\caption{
Rollout example where the baseline policy fails to place the green cube and then gets stuck trying to regrasp until timeout.
In comparison, RankQ (0.4) also fails to place the green cube initially but retries quickly after, succeeding before timeout. 
}
\label{fig:sim2real_snapshot_1}
\end{figure}

For the sim-to-real cube stacking experiments, we use an Interbotix WidowX 250S 7-DOF manipulator along with a single externally mounted Logitech C922 webcam.
We use PD control with gravity compensation to minimize tracking error.
The models are run on an NVIDIA RTX 4090 GPU after being optimized via \texttt{torch.compile}.
To reduce failures caused by dropping the cube from a height, we use more conservative binary gripper opening thresholds of 0.65 for RankQ (0.4) and 0.90 for RankQ (1.0). 
Applying similar conservative thresholds to Cal-QL and the initial VLA policy reduced performance, so we retain the default threshold of 0.50 for those methods.
Finally, sim-to-real rollouts compared with the baseline imitation policy can be seen in Figs.~\ref{fig:sim2real_snapshot_0} and \ref{fig:sim2real_snapshot_1}.

\newpage

\section{Tabulated Results}

\begin{table}[h]
\renewcommand{\arraystretch}{1.1}
\caption{Results for \texttt{antmaze-medium} environments (Fig.~\ref{fig:d4rl_results}). OSR refers to the success rate after offline RL.}
\centering
\footnotesize
\begin{tabular*}{0.9\textwidth}{@{\extracolsep{\fill}} l c c c c}
\toprule
\multirow{2}{*}[-3pt]{Method}
& \multicolumn{2}{c}{\texttt{medium-play}}
& \multicolumn{2}{c}{\texttt{medium-diverse}} \\
\cmidrule(lr){2-3}
\cmidrule(lr){4-5}
& OSR [$\uparrow$] & SR [$\uparrow$] & OSR [$\uparrow$] & SR [$\uparrow$] \\
\midrule
CQL & \meanstd{\mathbf{0.817}}{0.047} & \meanstd{0.980}{0.005} & \meanstd{0.667}{0.085} & \meanstd{\uline{0.981}}{0.003} \\
CQL+SAC & \meanstd{\uline{0.783}}{0.085} & \meanstd{0.983}{0.005} & \meanstd{\uline{0.783}}{0.094} & \meanstd{\mathbf{0.983}}{0.004} \\
Cal-QL & \meanstd{0.717}{0.062} & \meanstd{0.978}{0.009} & \meanstd{\mathbf{0.817}}{0.118} & \meanstd{0.958}{0.006} \\
Cal-QL+SAC & \meanstd{0.750}{0.108} & \meanstd{\mathbf{0.989}}{0.002} & \meanstd{0.717}{0.062} & \meanstd{0.969}{0.010} \\
Hybrid RL & \meanstd{0.133}{0.155} & \meanstd{0.983}{0.006} & \meanstd{0.017}{0.024} & \meanstd{0.971}{0.010} \\
SAC & \meanstd{0.000}{0.000} & \meanstd{0.000}{0.000} & \meanstd{0.000}{0.000} & \meanstd{0.000}{0.000} \\
SAC+OFF & \meanstd{0.117}{0.165} & \meanstd{\uline{0.988}}{0.006} & \meanstd{0.117}{0.062} & \meanstd{0.966}{0.015} \\
RankQ & \meanstd{0.750}{0.147} & \meanstd{0.977}{0.006} & \meanstd{0.667}{0.165} & \meanstd{0.962}{0.023} \\
RankQ+SAC & \meanstd{\mathbf{0.817}}{0.085} & \meanstd{0.987}{0.001} & \meanstd{\uline{0.783}}{0.131} & \meanstd{0.967}{0.011} \\
\bottomrule
\end{tabular*}
\label{tab:antmaze-medium}
\end{table}

\begin{table}[h]
\renewcommand{\arraystretch}{1.1}
\caption{Results for \texttt{antmaze-large} environments. OSR refers to the success rate after offline RL.}
\centering
\footnotesize
\begin{tabular*}{0.9\textwidth}{@{\extracolsep{\fill}} l c c c c}
\toprule
\multirow{2}{*}[-3pt]{Method}
& \multicolumn{2}{c}{\texttt{large-play}}
& \multicolumn{2}{c}{\texttt{large-diverse}} \\
\cmidrule(lr){2-3}
\cmidrule(lr){4-5}
& OSR [$\uparrow$] & SR [$\uparrow$] & OSR [$\uparrow$] & SR [$\uparrow$] \\
\midrule
CQL & \meanstd{0.283}{0.094} & \meanstd{\uline{0.828}}{0.032} & \meanstd{0.100}{0.071} & \meanstd{0.210}{0.261} \\
CQL+SAC & \meanstd{\uline{0.433}}{0.047} & \meanstd{0.000}{0.000} & \meanstd{\uline{0.233}}{0.184} & \meanstd{0.001}{0.001} \\
Cal-QL & \meanstd{0.367}{0.125} & \meanstd{0.677}{0.193} & \meanstd{0.183}{0.085} & \meanstd{\uline{0.740}}{0.053} \\
Cal-QL+SAC & \meanstd{0.300}{0.108} & \meanstd{0.000}{0.000} & \meanstd{\mathbf{0.250}}{0.141} & \meanstd{0.000}{0.000} \\
Hybrid RL & \meanstd{0.000}{0.000} & \meanstd{0.000}{0.000} & \meanstd{0.000}{0.000} & \meanstd{0.000}{0.000} \\
SAC & \meanstd{0.000}{0.000} & \meanstd{0.000}{0.000} & \meanstd{0.000}{0.000} & \meanstd{0.000}{0.000} \\
SAC+OFF & \meanstd{0.000}{0.000} & \meanstd{0.000}{0.000} & \meanstd{0.000}{0.000} & \meanstd{0.001}{0.001} \\
RankQ & \meanstd{0.367}{0.143} & \meanstd{\mathbf{0.912}}{0.056} & \meanstd{0.217}{0.131} & \meanstd{\mathbf{0.847}}{0.018} \\
RankQ+SAC & \meanstd{\mathbf{0.467}}{0.131} & \meanstd{0.000}{0.000} & \meanstd{0.217}{0.085} & \meanstd{0.000}{0.000} \\
\bottomrule
\end{tabular*}
\label{tab:antmaze-large}
\end{table}

\begin{table}[h]
\renewcommand{\arraystretch}{1.1}
\caption{Results for \texttt{adroit} environments. OSR refers to the success rate after offline RL.}
\centering
\scriptsize
\begin{tabular*}{\textwidth}{@{\extracolsep{\fill}} l c c c c c c}
\toprule
\multirow{2}{*}[-3pt]{Method}
& \multicolumn{2}{c}{\texttt{pen}}
& \multicolumn{2}{c}{\texttt{door}}
& \multicolumn{2}{c}{\texttt{relocate}} \\
\cmidrule(lr){2-3}
\cmidrule(lr){4-5}
\cmidrule(lr){6-7}
& OSR [$\uparrow$] & SR [$\uparrow$] & OSR [$\uparrow$] & SR [$\uparrow$] & OSR [$\uparrow$] & SR [$\uparrow$] \\
\midrule
CQL & \meanstdtiny{\uline{0.533}}{0.062} & \meanstdtiny{0.816}{0.035} & \meanstdtiny{\mathbf{0.067}}{0.024} & \meanstdtiny{0.917}{0.085} & \meanstdtiny{0.000}{0.000} & \meanstdtiny{0.632}{0.126} \\
CQL+SAC & \meanstdtiny{\uline{0.533}}{0.062} & \meanstdtiny{0.711}{0.063} & \meanstdtiny{\uline{0.033}}{0.024} & \meanstdtiny{\mathbf{0.991}}{0.011} & \meanstdtiny{0.000}{0.000} & \meanstdtiny{0.774}{0.004} \\
Cal-QL & \meanstdtiny{\mathbf{0.750}}{0.082} & \meanstdtiny{\mathbf{0.987}}{0.004} & \meanstdtiny{\mathbf{0.067}}{0.062} & \meanstdtiny{0.932}{0.090} & \meanstdtiny{0.000}{0.000} & \meanstdtiny{\mathbf{0.937}}{0.045} \\
Cal-QL+SAC & \meanstdtiny{\mathbf{0.750}}{0.082} & \meanstdtiny{0.833}{0.090} & \meanstdtiny{\uline{0.033}}{0.047} & \meanstdtiny{0.982}{0.015} & \meanstdtiny{0.000}{0.000} & \meanstdtiny{0.737}{0.050} \\
Hybrid RL & \meanstdtiny{0.033}{0.047} & \meanstdtiny{0.738}{0.046} & \meanstdtiny{0.000}{0.000} & \meanstdtiny{\uline{0.989}}{0.008} & \meanstdtiny{0.000}{0.000} & \meanstdtiny{0.737}{0.064} \\
SAC & \meanstdtiny{0.133}{0.085} & \meanstdtiny{0.488}{0.051} & \meanstdtiny{0.000}{0.000} & \meanstdtiny{0.489}{0.401} & \meanstdtiny{0.000}{0.000} & \meanstdtiny{0.004}{0.002} \\
SAC+OFF & \meanstdtiny{0.033}{0.047} & \meanstdtiny{0.737}{0.020} & \meanstdtiny{0.000}{0.000} & \meanstdtiny{0.953}{0.024} & \meanstdtiny{0.000}{0.000} & \meanstdtiny{0.443}{0.110} \\
RankQ & \meanstdtiny{0.050}{0.041} & \meanstdtiny{\uline{0.969}}{0.023} & \meanstdtiny{0.000}{0.000} & \meanstdtiny{\mathbf{0.991}}{0.002} & \meanstdtiny{0.000}{0.000} & \meanstdtiny{\uline{0.932}}{0.006} \\
RankQ+SAC & \meanstdtiny{0.050}{0.041} & \meanstdtiny{0.912}{0.045} & \meanstdtiny{0.000}{0.000} & \meanstdtiny{0.941}{0.042} & \meanstdtiny{0.000}{0.000} & \meanstdtiny{0.817}{0.028} \\
\bottomrule
\end{tabular*}
\label{tab:adroit}
\end{table}

\newpage

\begin{table}[h]
\renewcommand{\arraystretch}{1.1}
\caption{Table results for \texttt{vla-low-data} environments (Fig.~\ref{fig:vla-low-data_results}). OSR refers to the success rate after offline RL. The VLA's initial success rate is shown in parentheses for each environment.}
\centering
\scriptsize
\begin{tabular*}{\textwidth}{@{\extracolsep{\fill}}  l c c c c c c}
\toprule
\multirow{2}{*}[-3pt]{Method} & \multicolumn{2}{c}{\texttt{carrot-onto-plate} $(0.735)$} & \multicolumn{2}{c}{\texttt{cube-stacking} $(0.240)$} & \multicolumn{2}{c}{\texttt{spoon-into-bowl} $(0.205)$} \\
\cmidrule(lr){2-3}
\cmidrule(lr){4-5}
\cmidrule(lr){6-7}
& OSR [$\uparrow$] & SR [$\uparrow$] & OSR [$\uparrow$] & SR [$\uparrow$] & OSR [$\uparrow$] & SR [$\uparrow$] \\
\midrule
CQL & \meanstdtiny{0.743}{0.071}  & \meanstdtiny{0.810}{0.056}  & \meanstdtiny{0.078}{0.029} & \meanstdtiny{0.260}{0.083} & \meanstdtiny{0.167}{0.119} & \meanstdtiny{0.258}{0.016}  \\
CQL+SAC & \meanstdtiny{0.675}{0.103}  & \meanstdtiny{\uline{0.835}}{0.027}  & \meanstdtiny{0.047}{0.045} & \meanstdtiny{0.047}{0.020} & \meanstdtiny{0.118}{0.084} & \meanstdtiny{0.232}{0.090} \\
Cal-QL & \meanstdtiny{0.330}{0.214}  & \meanstdtiny{0.737}{0.118}  & \meanstdtiny{0.088}{0.068} & \meanstdtiny{0.297}{0.073} & \meanstdtiny{0.053}{0.075} & \meanstdtiny{0.208}{0.111}  \\
Cal-QL+SAC & \meanstdtiny{0.463}{0.261}  & \meanstdtiny{0.775}{0.023}  & \meanstdtiny{0.083}{0.043} & \meanstdtiny{0.018}{0.019} & \meanstdtiny{0.143}{0.098} & \meanstdtiny{0.207}{0.067} \\
SAC+OFF & \meanstdtiny{0.000}{0.000}  & \meanstdtiny{0.645}{0.029}  & \meanstdtiny{0.000}{0.000} & \meanstdtiny{0.147}{0.150} & \meanstdtiny{0.000}{0.000} & \meanstdtiny{0.153}{0.141} \\
RLPD & N/A & \meanstdtiny{0.420}{0.108} & N/A & \meanstdtiny{0.060}{0.065} & N/A & \meanstdtiny{0.003}{0.005} \\
WSRL & \meanstdtiny{0.583}{0.266} & \meanstdtiny{0.757}{0.085} & \meanstdtiny{0.287}{0.207} & \meanstdtiny{0.260}{0.226} & \meanstdtiny{0.093}{0.057} & \meanstdtiny{\uline{0.393}}{0.165} \\
RankQ & \meanstdtiny{\mathbf{0.897}}{0.047}  & \meanstdtiny{\mathbf{0.930}}{0.007}  & \meanstdtiny{\mathbf{0.707}}{0.172} & \meanstdtiny{\mathbf{0.940}}{0.018} & \meanstdtiny{\mathbf{0.478}}{0.124} & \meanstdtiny{\mathbf{0.800}}{0.065} \\
RankQ+SAC & \meanstdtiny{\uline{0.835}}{0.039}  & \meanstdtiny{0.585}{0.410}  & \meanstdtiny{\uline{0.597}}{0.218} & \meanstdtiny{\uline{0.328}}{0.464} & \meanstdtiny{\uline{0.355}}{0.168} & \meanstdtiny{0.085}{0.120} \\
\bottomrule
\end{tabular*}
\label{tab:vla-low-data}
\end{table}

\begin{table}[h]
\renewcommand{\arraystretch}{1.1}
\caption{Table results for the \texttt{vla-sim2real-cube-stacking} environment (Fig.~\ref{fig:vla-sim2real_results}). 
OTSR refers to the training rollout success rate after offline RL. 
TSR refers to the training rollout success rate after online training. 
TTF refers to the average time-to-finish for successful training rollouts. 
Note that training rollout metrics are slightly lower than evaluation metrics due to stochastic action sampling during training rather than greedy mode selection during evaluation.
The VLA's initial success rate and TTF are shown in parentheses.}
\centering
\footnotesize
\begin{tabular}{l c c c}
\toprule
\multirow{2}{*}[-3pt]{Method} & \multicolumn{3}{c}{\texttt{cube-stacking} $(0.080, 12.53)$} \\
\cmidrule(lr){2-4}
& OTSR [$\uparrow$] & TSR [$\uparrow$] & TTF (s) [$\downarrow$]  \\
\midrule
CQL & \meanstd{\uline{0.056}}{0.052} & \meanstd{0.690}{0.066} & \meanstd{11.42}{0.53} \\
Cal-QL & \meanstd{\mathbf{0.111}}{0.079} & \meanstd{\uline{0.743}}{0.054} & \meanstd{\uline{10.51}}{0.24} \\
SAC+OFF & \meanstd{0.014}{0.020} & \meanstd{0.128}{0.181} & - \\
RankQ & \meanstd{0.042}{0.034} & \meanstd{\mathbf{0.880}}{0.016} & \meanstd{\mathbf{8.36}}{0.01} \\
\bottomrule
\end{tabular}
\label{tab:vla-sim2real}
\end{table}

\newpage

\end{document}